\documentclass[letterpaper,twocolumn,10pt]{article}
\usepackage{usenix}

\usepackage{tikz}
\usepackage{amsmath}
\usepackage{enumitem}
\usepackage{url}
\usepackage{microtype}
\usepackage{multirow}
\usepackage[skip=2pt]{caption}
\usepackage{booktabs}
\usepackage{amsmath,amsfonts}
\newtheorem{definition}{Definition}

\begin{document}

\date{}

\title{
\Large \bf How Generalizable are Deepfake Image Detectors? An Empirical Study
\thanks{Boquan Li, Jun Sun and Christopher M. Poskitt are with School of Computing and Information Systems, Singapore Management University.
Boquan Li and Xingmei Wang are with College of Computer Science and Technology, Harbin Engineering University.
Contact at liboquan@hrbeu.edu.cn.}
}
\author{\rm Boquan Li, Jun Sun, Christopher M. Poskitt, and Xingmei Wang}

\maketitle

\begin{abstract}
Deepfakes are becoming increasingly credible, posing a significant threat given their potential to facilitate fraud or bypass access control systems.
This has motivated the development of deepfake detection methods, in which deep learning models are trained to distinguish between real and synthesized footage.
Unfortunately, existing detectors struggle to generalize to deepfakes from datasets they were not trained on, but little work has been done to examine why or how this limitation can be addressed.
Especially, those single-modality deepfake images reveal little available forgery evidence, posing greater challenges than detecting deepfake videos.
In this work, we present the first empirical study on the generalizability of deepfake detectors, an essential goal for detectors to stay one step ahead of attackers.
Our study utilizes six deepfake datasets, five deepfake image detection methods, and two model augmentation approaches, confirming that detectors do not generalize in zero-shot settings.
Additionally, we find that detectors are learning unwanted properties specific to synthesis methods and struggling to extract discriminative features, limiting their ability to generalize.
Finally, we find that there are neurons universally contributing to detection across seen and unseen datasets, suggesting a possible path towards zero-shot generalizability.
\end{abstract}

\section{Introduction}

Deepfakes are images or videos manipulated using deep learning algorithms~\cite{tolosana2020deepfakes}.
The technology behind them has advanced to such a state that the best deepfakes are nearly indistinguishable from authentic footage.
As deepfakes become more accessible and easier to create, they pose an increasing threat to individuals, organizations and governments, given their potential use for spreading misinformation, manipulating public opinion, committing fraud, or even bypassing access control systems~\cite{li2022seeing}.

In response to this growing risk, a considerable amount of research has been conducted on methods for detecting deepfake videos and images.
The latest deepfake method, \texttt{Wav2Lip}~\cite{khalid2021fakeavceleb}, utilizes speech content to guide the sync of lip movements, generating high-reality deepfake videos.
Fortunately, existing efforts report that the correlative visual and audio features have left available evidence for deepfake detection, e.g., by capturing the temporal synchronization between visual and audio frames in videos~\cite{feng2023self}.
By contrast, the single-modality deepfake images leave limited forgery evidence, posing greater challenges than detecting deepfake videos, and are thus the main focus of this work.

Existing deepfake image detection methods can be categorized into two groups.
The first group includes methods that rely on extracting visual features, such as visual artifacts in eyes and teeth~\cite{matern2019exploiting}, or facial surrounding regions~\cite{li2018exposing}.
However, these detectors are impractical in general, as such features are susceptible to objective factors such as environmental conditions or image compression, and can be avoided by newer deepfake synthesis methods~\cite{li2020celeb}.
The second group includes methods based on deep learning, such as \texttt{MesoNet}\cite{afchar2018mesonet} and \texttt{XceptionNet}\cite{chollet2017xception}, which classify real and deepfake images based on their microscopic differences.
These methods overcome the practical limitations of visual feature-specific detectors.
However, given the rapid development of deepfake technology, it is imperative that deepfake detectors are \emph{generalizable}, meaning that their effectiveness can be maintained across different deepfake datasets.

Research efforts have attempted to make deepfake image detectors generalizable, and can be broadly categorized into two groups.
The first group of methods~\cite{cozzolino2018forensictransfer,nguyen2019multi,kim2021cored,lee2021tar,kim2021fre} achieves generalizable deepfake detection in \emph{few-shot} settings, where techniques like transfer learning are applied to a few \emph{unseen} deepfakes (i.e., samples from a different deepfake dataset than that initially trained on).
While these methods report promising results, they will always be one step behind new deepfake datasets given that unseen samples are required for generalization.
The second group of methods~\cite{sun2021domain,du2020towards,chen2022self,tariq2021one,xu2022supervised}, while primarily concerned with few-shot settings, also make tentative attempts in \emph{zero-shot} settings, i.e., without requiring unseen deepfakes from other datasets.
While the zero-shot setting is the most practical, unfortunately, these methods have only limited success in it.
Specifically, their results are only reported on certain unseen datasets~\cite{sun2021domain,du2020towards,chen2022self,tariq2021one,xu2022supervised}, show limited effectiveness~\cite{sun2021domain,du2020towards}, or are even contradicted in different research~\cite{chen2022self,tariq2021one}.
This suggests that existing deepfake image detection efforts are not generalizable.

While it is intuitively clear that we have not yet reached generalizable deepfake image detection, we lack a systematic, comparative evaluation of the state-of-art.
It would be useful to know which combinations of deepfake image detectors and model augmentation approaches lead to better generalizability.
Additionally, it would be advantageous to gain insights into why deepfake image detectors fail to generalize.
For example, is it possible to extract information from their models to explain why zero-shot generalizability is challenging to achieve?
Are there learned features in the models that would be useful for achieving generalizability?
To our knowledge, no empirical research has been undertaken to explore these factors in the context of deepfake detection.

To address this, we conduct an empirical study to investigate the generalizability of deepfake image detectors.
In particular, our research questions and findings include:

\begin{itemize}
	\item
	To what extent do current deepfake image detectors generalize? Is it true that state-of-the-art detectors generalize poorly? Through this question,
	Our study confirms that existing detectors are not generalizable in a zero-shot setting (with some mitigation possible in a few-shot one).
	\item
	Assuming detectors truly lack generalizability, are they learning `unwanted' properties of deepfakes that fail to characterize the essential differences between real and fake images?
	Through this question, we find that detectors are learning unwanted properties specific to synthesis methods that interfere with their ability to generalize.
	\item
	Using interpretable artificial intelligence~(AI) techniques, can we determine whether any of the features that detectors do extract are discriminative enough to classify real and fake images?
	Through this question, we find that detectors do not extract discriminative features that support generalizability.
	\item
	We pivot away from extracted features and investigate detection models on a more granular level: in particular, are there neurons in the underlying neural networks that are contributing to the detection of deepfakes across both seen and unseen datasets?
	Through this question, we find the existence of such universal neurons, which suggests a possible new focus of future work aiming for generalizable zero-shot deepfake detection.
\end{itemize}

Furthermore, we provide a repository of detectors and datasets resulting from our study\footnote{We release our datasets, detectors and codes in a
Github repository. \url{https://github.com/boutiquelee/DeepfakeEmpiricalStudy}}, which can serve as baselines or foundational blocks to support future research and applications.

The remainder of this paper is organized as follows.
We introduce the background and some key related work to this study in Section~\ref{relatedwork}.
Next, we present the design and systematical experiments respectively in Section~\ref{design} and Section~\ref{study}.
Finally, we discuss the potential research directions as well as some threats to validity in Section~\ref{discussion} and Section~\ref{threats} before concluding in Section~\ref{conculsion}.

\section{Background and Related Work}
\label{relatedwork}

In this section, we review the necessary background on deepfake generation, deepfake image detection, and the generalizability of detectors.
Following this, we compare our work against some related existing empirical studies on deepfakes.

\subsection{Deepfake Generation}
\label{generation}

\begin{figure}[t]
	\centering
	\includegraphics[width=0.75\linewidth]{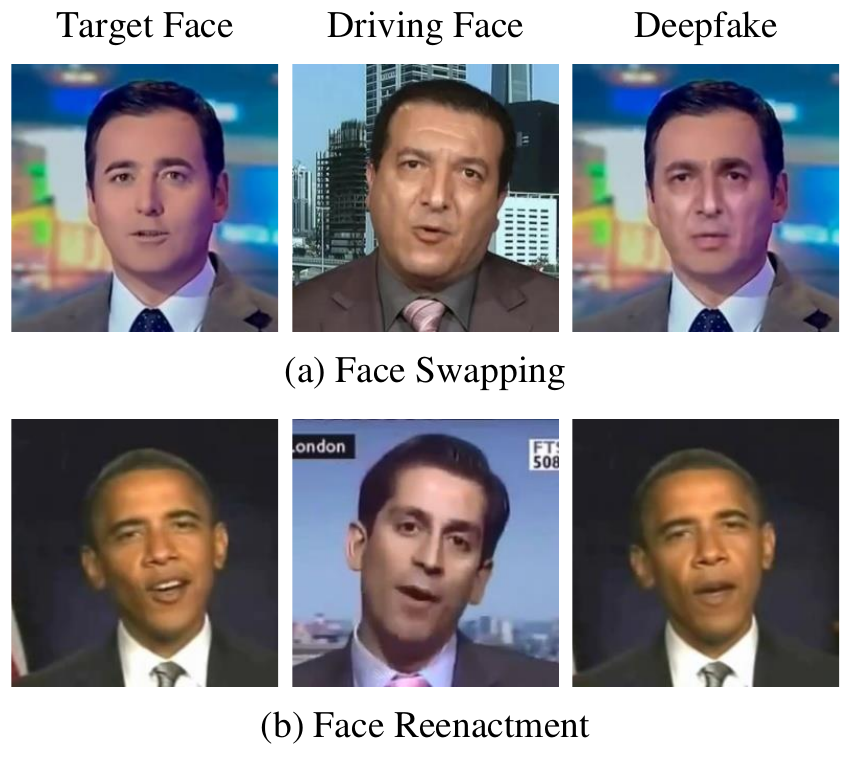}
	\caption{Comparison of deepfake generation methods.}
	\label{fig:example}
\end{figure}

A deepfake is a synthetic image (or video) in which the face of a \emph{target} individual is manipulated according to the face or expressions of a \emph{driving face}.
Current deepfake generation methods can be classified as performing either \emph{face swapping} or \emph{face reenactment}.
In face swapping, the target provides the facial motions and expressions, but the face itself is replaced by that of the driving face.
In face reenactment, however, the target retains his/her face but the motions are transformed into the ones provided by the driving face.
Figure~\ref{fig:example}~\cite{rossler2019faceforensics++} illustrates the difference between these two categories of methods.

\begin{table*}[t]
	\caption{State-of-the-art deepfake generation methods.}
	\begin{center}
		\tiny
		\renewcommand\arraystretch{1.2}
		\resizebox{\linewidth}{!}
		{
			\begin{tabular}{|c|c|c|c|}
				\hline
				\textbf{Method}            
				& \textbf{Category}         
				& \textbf{Algorithm}           
				& \textbf{Project Page}                                                                
				\\ \hline
				\texttt{FaceSwap}    
				& Face Swapping    
				& Computer graphics   
				& \url{https://github.com/MarekKowalski/FaceSwap}
				\\ \hline
				\texttt{DeepFake}                                                
				& Face Swapping    
				& AutoEncoder         
				& \url{https://github.com/deepfakes/faceswap}
				\\ \hline
				\texttt{FaceSwap-GAN}  
				& Face Swapping    
				& AutoEncoder+GAN     
				& \url{https://github.com/shaoanlu/faceswap-GAN}
				\\ \hline
				\texttt{Face2Face}~\cite{thies2016face2face}        
				& Face Reenactment 
				& Computer graphics   
				& \url{http://niessnerlab.org/projects/thies2016face.html}
				\\ \hline
				\texttt{Neural Textures}~\cite{thies2019deferred}
				& Face Reenactment 
				& GAN+Neural Textures
				& \url{http://niessnerlab.org/projects/thies2019neural.html}
				\\ \hline
			\end{tabular}
		}\label{tab:dfmethod}
	\end{center}
\end{table*}

Table~\ref{tab:dfmethod} summarizes the state-of-the-art deepfake generation methods across these two categories.
The methods rely on a variety of underlying algorithms from image recognition and computer graphics (e.g., 3D model fitting), as well as deep neural networks~(DNNs) such as autoencoders~\cite{badrinarayanan2017segnet} and generative adversarial networks~(GANs)~\cite{goodfellow2014generative}.

\textbf{Study Scope.}
In addition to the above deepfake methods, the latest \texttt{Wav2Lip}~\cite{khalid2021fakeavceleb} utilizes speech content to guide the sync of lip movements, generating high-reality deepfakes.
In response to \texttt{Wav2Lip}, existing research~\cite{feng2023self} reports that such correlative visual and audio information in videos leaves available evidence for deepfake detection, indicating effective solutions to capture the temporal synchronization between visual and audio frames.
By comparison, the single-modality deepfake images leave limited forgery evidence, which poses greater challenges than detecting deepfake videos.
Thus, in this study, we focus on the generalizability of deepfake image detectors, which still have plenty of room to be investigated.

Additionally, the key threat of deepfake generation arises from the replacement of one human identity with another: not only can this lead to social and political consequences, but there are also security implications, e.g., in bypassing face-based access control systems~\cite{li2022seeing}.
Note that technologies for synthesizing `new' or non-existent object or human images, such as stable diffusion~\cite{rombach2022high} or image translation~\cite{isola2017image} algorithms, are considered out of the scope of our study, as the aforementioned threats only apply when considering the faces (i.e., identities) of real humans.

\subsection{Deepfake Image Detection}
\label{sec:detector}

Deepfake image detection methods generally fall into one of two categories, based on \emph{visual features} or \emph{deep learning}.
Table~\ref{tab:detector} summarizes the state-of-the-art methods across these two categories, which we expand upon in the following.

\begin{table*}[t]
	\caption{State-of-the-art deepfake image detection methods.}
	\begin{center}
		\Huge
		\renewcommand\arraystretch{1.2}
		\resizebox{\linewidth}{!}
		{
			\begin{tabular}{|c|c|c|c|c|}
				\hline
				\textbf{Detector}      
				& \textbf{Category}        
				& \textbf{Feature}                            
				& \textbf{Classifier}      
				& \textbf{Project Page}
                \\ \hline
				\texttt{VA-MLP}~\cite{matern2019exploiting}  
				& Visual features
				& Eyes, teeth and facial contours    
				& MLP and LogReg  
				& Unpublished                                   \\ \hline
				\texttt{HeadPose}~\cite{yang2019exposing}     
				& Visual features
				& Head poses                         
				& SVM             
				& \url{https://bitbucket.org/ericyang3721/headpose\_forensic}  
                \\ \hline
				\texttt{FWA}~\cite{li2018exposing}   
				& Visual features
				& Face areas and surrounding regions 
				& CNN             
				& \url{https://github.com/yuezunli/CVPRW2019\_Face\_Artifacts} 
                \\ \hline
				\texttt{WorldLeader}~\cite{agarwal2019protecting}  
				& Visual features
				& Facial expressions   
				& SVM             
				& Unpublished   
				\\ \hline
				\texttt{MesoNet}~\cite{afchar2018mesonet}     
				& Deep learning   
				& Microscopic-level features                
				& CNN             
				& \multirow{2}{*}{\url{https://github.com/DariusAf/MesoNet}}   \\ \cline{1-4}
				\texttt{MesoInception}~\cite{afchar2018mesonet}
				& Deep learning   
				& Microscopic-level features
				& CNN             
				& \\ \hline
				\texttt{ShallowNet}~\cite{tariq2019gan}    
				& Deep learning   
				& Microscopic-level features
				& CNN             
				& \url{https://github.com/shahroztariq/ShallowNet}             
                \\ \hline
				\texttt{XceptionNet}~\cite{chollet2017xception} 
				& Deep learning   
				& Microscopic-level features   
				& CNN             
				& \url{https://github.com/ondyari/FaceForensics}               
                \\ \hline
				\texttt{EfficientNet}~\cite{tan2019efficientnet}
				& Deep learning   
				& Microscopic-level features                    
				& CNN             
				& \url{https://github.com/aaronchong888/DeepFake-Detect}       
                \\ \hline
			\end{tabular}
		}\label{tab:detector}
	\end{center}
\end{table*}

\textbf{Visual Feature Detectors.}
The first category includes techniques that extract and analyze visual features, e.g., artifacts, for inconsistencies or search for abnormalities arising from deepfake generation.
For example, Matern et al.~\cite{matern2019exploiting} explored reflection details in eyes, teeth, and facial contours in deepfakes, and Yang et al.~\cite{yang2019exposing} extracted a full set of facial landmarks to estimate 3D head poses and detect abnormal poses.
Similarly,
Li et al.~\cite{li2018exposing} explored the face-warping step in deepfake creation and extracted the visual artifacts in face areas and their surrounding regions for classification.
Agarwal et al.~\cite{agarwal2019protecting} focused on specific people and extracted facial expressions and movements to detect deepfakes based on identity authentication.
However, these detectors have limitations in practical applications since specific visual features are susceptible to environmental conditions (such as lights) or image compression, and can be avoided by newer deepfake methods~\cite{li2020celeb}.

\textbf{Deep Learning Detectors.}
The second category utilizes DNNs such as convolutional neural networks~(CNNs)~\cite{li2021survey} to classify real and deepfake images based on microscopic-level features (i.e., pixel-level differences). 
For example, Afchar et al.~\cite{afchar2018mesonet} proposed \texttt{MesoNet}, an efficient deepfake detector based on a lightweight network architecture.
They also devised the improved \texttt{MesoInception} by appending a variant of the Inception module~\cite{szegedy2015going} to \texttt{MesoNet}.
Tariq et al.~\cite{tariq2019gan} proposed three versions of \texttt{ShallowNet}, a method for detecting GAN-based deepfakes, among which \texttt{ShallowNet-V3} achieved the best performance.
Additionally, Rossler et al.~\cite{rossler2019faceforensics++} and \cite{dfdetect2022} utilized the advanced \texttt{XceptionNet}~\cite{chollet2017xception} and \texttt{EfficientNet}~\cite{tan2019efficientnet} CNNs as deepfake detectors, both of which achieved promising performance.
Compared to visual feature-based methods, deep learning detectors are more practical since they focus on microscopic-level features (e.g., pixels) for classification, overcoming the limitations of feature-specific solutions.

\subsection{Generalizability of Deepfake Image Detectors}
\label{sec:generalizability}

The deepfake detection methods we have discussed thus far typically focus on identifying deepfakes generated by one particular dataset.
For detection methods to be practical in general, and for them to stay one step ahead of attackers, they must be able to generalize to identifying any upcoming deepfakes.
Existing approaches towards this goal can be categorized as either \emph{few-shot} or \emph{zero-shot}: those in the former category require some additional deepfake examples from unseen datasets, whereas those in the latter category do not.
Table~\ref{tab:generalization} summarizes the state-of-the-art generalization methods within these two categories, which we expand upon in the following.

\begin{table*}[!t]
    \caption{State-of-the-art generalization methods for deepfake image detection.}
	\begin{center}
        \LARGE
		\renewcommand\arraystretch{1.2}
		\resizebox{\linewidth}{!}
		{
			\begin{tabular}{|c|c|c|c|}
				\hline
				\textbf{Method}                    
				& \textbf{Category}          
				& \textbf{Augmentation Strategy}
				& \textbf{Project Page}                                       \\ \hline
				\texttt{Forensictransfer}~\cite{cozzolino2018forensictransfer}
				& Few-shot  
				& Transfer Learning                                    
				& Unpublished                                        \\ \hline
				\texttt{ClassNSeg}~\cite{nguyen2019multi}       
				& Few-shot  
				& Multi-task Learning                                  
				& \url{https://github.com/nii-yamagishilab/ClassNSeg}      \\ \hline
				\texttt{CoReD}~\cite{kim2021cored}            
				& Few-shot
				& Continual Representation Learning                  
				& \url{https://github.com/alsgkals2/CoReD\_Released}       \\ \hline
				\texttt{FReTAL}~\cite{kim2021fre}       
				& Few-shot
				& Knowledge Distillation and Representation Learning 
				& \url{https://github.com/simrit1/FReTAL}                  \\ \hline
				\texttt{TAR}~\cite{lee2021tar}     
				& Few-shot 
				& Weakly-supervised Learning                         
				& \url{https://github.com/Clench/TAR\_resAE}               \\ \hline
				\texttt{LTW}~\cite{sun2021domain}         
				& Few-shot and Zero-shot
				& Meta-weight Learning                                 
				& \url{https://github.com/skJack/LTW}                      \\ \hline
				\texttt{LAE}~\cite{du2020towards}             
				& Few-shot and Zero-shot
				& Active Learning     
				& Unpublished                                        \\ \hline
				\texttt{SLADD}~\cite{chen2022self}           
				& Few-shot and Zero-shot
				& Adversarial Augmentation                                    
				& \url{https://github.com/liangchen527/SLADD}              \\ \hline
				\texttt{OneRuleAll}~\cite{tariq2021one}     
				& Few-shot and Zero-shot
				& Merge Learning and Transfer Learning               
				& \url{https://github.com/shahroztariq/CLRNet}             \\ \hline
				\texttt{SupCon}~\cite{xu2022supervised}          
				& Few-shot and Zero-shot
				& Supervised Contrastive Learning                    
				& \url{https://github.com/xuyingzhongguo/deepfake\_supcon} \\ \hline
			\end{tabular}
		}\label{tab:generalization}
	\end{center}
\end{table*}

\textbf{Few-Shot Generalization.}
Methods in this category take deepfake detectors trained on one dataset, and generalize them by providing a few additional training examples from an unseen one.
Cozzolino et al.~\cite{cozzolino2018forensictransfer} introduced a transfer learning approach that fine-tunes a detector to learn a forensic embedding for deepfake detection.
Nguyen et al.~\cite{nguyen2019multi} proposed a multi-task learning strategy that fine-tunes detectors in semi-supervised settings.
Kim et al.~\cite{kim2021cored} proposed a continual representation learning approach that performs domain adaptation between seen and unseen deepfakes.
Similarly, Kim et al.~\cite{kim2021fre} used both knowledge distillation and representation learning strategies to perform domain adaptation on unseen deepfakes while minimizing `catastrophic forgetting' (losing the model's prior knowledge about deepfakes).
Lee et al.~\cite{lee2021tar} developed an autoencoder with residual blocks that perform weakly-supervised training on deepfake detectors.
While these methods may achieve some generalization, they may not be practical in practice, since unseen deepfakes are often unavailable.
Furthermore, it means that the detection method will always be one step behind an attacker that deploys a novel approach to deepfake generation.

\textbf{Zero-Shot Generalization.}
As shown in Table~\ref{tab:generalization}, existing works have predominantly focused on few-shot generalization.
However, some authors have taken tentative steps to apply their methods in a zero-shot setting as well, i.e., in which (generalizable) deepfake image detectors are trained without requiring examples from unseen datasets.
In theory, zero-shot generalization is more practical, but existing efforts that reported promising results in few-shot settings, unfortunately, only showed limited success in zero-shot ones.

In some research, results have only been reported on certain datasets or have only shown limited effectiveness.
Sun et al.~\cite{sun2021domain} proposed a meta-weight learning strategy that performed domain adaptation between different datasets and achieved an accuracy of about 63\% on unseen deepfakes.
Du et al.~\cite{du2020towards} proposed a locality-aware autoencoder and an active learning framework, which achieved an accuracy of about 68\% on a specific unseen dataset.
Chen et al.~\cite{chen2022self} proposed an adversarial augmentation strategy, which achieved a passable performance, again, on a specific unseen dataset.

In other research, contradicting results have been reported.
Tariq et al.~\cite{tariq2021one} used merge learning and transfer learning strategies to fine-tune baseline detectors by augmenting multiple types of seen deepfakes, resulting in passable performance on specific unseen datasets.
However, Xu et al.~\cite{xu2022supervised} reported that this strategy did not work on additional unseen datasets.
To address this issue, they proposed a supervised contrastive loss strategy, which reported passable results on a specific dataset but showed poor performance on other ones.

\subsection{Empirical Studies on Deepfakes}

From the aforementioned work we have surveyed, it is intuitively clear that we have not yet reached (zero-shot) generalizable deepfake image detection.
We lack, however, a systematic evaluation of the state-of-the-art with respect to this goal.
It would be useful to systemize `\emph{where we are}' towards achieving generalizable methods, why current methods fall short, and how generalizability might be achieved, which are the motivations of this empirical study.
While (to the best of our knowledge) this is the first such empirical study on the topic, there have been a number of other empirical studies on other topics related to deepfakes.

Several empirical studies have explored the impact of deepfakes in sociological domains.
Vaccari et al.~\cite{vaccari2020deepfakes} investigated the contribution of deepfakes to online disinformation and highlighted the challenge they pose to online civic culture.
Gamage et al.~\cite{gamage2022deepfakes} examined the social implications of deepfakes and proposed prospective management measures to mitigate their potential societal harms.
Appel et al.~\cite{appel2022detection} focused on political deepfakes and investigated the analytic thinking and political interests of citizens, as predictors of correctly detecting deepfakes.
Our empirical study differs by focusing on the generalizability of deepfake image detectors within technical domains only.

Other empirical studies have focused on the generation of non-existent objects or humans.
Zhao et al.~\cite{zhao2021deep} studied the algorithmic mechanism of falsifying satellite images with non-existent landscape features.
Liu et al.~\cite{liu2020global} studied entirely non-existent faces and investigated the transferability of GAN fingerprints for their detection.
Yang et al.~\cite{yang2022deepfake} also focused on non-existent faces and studied the network architecture attribution for detection.
In contrast to these studies, our work specifically focuses on deepfakes that involve human identity replacements, such as face swapping and face reenactment, which pose actual threats to information security.

\section{Study Design}
\label{design}

In this section, we present the design of our empirical study, including our research questions, datasets, baseline detectors, and implementation.

\subsection{Research Questions}
\label{sec:research_questions}

The overall goal of our study is to experimentally assess the extent to which state-of-the-art deepfake image detectors can generalize, why current methods fail to generalize, and how generalizability might be achieved.
In particular, we aim to answer the following research questions~(RQs):

\begin{itemize}
	\item
	\textit{RQ1: Generalizability.}
	Do existing detectors generalize to deepfakes across different datasets?
	\item
	\textit{RQ2: Unwanted Properties.}
	Do detectors learn deepfake properties that are unwanted for generalization?
	\item
	\textit{RQ3: Discriminative Features.}
	Do detectors extract meaningful features that are discriminative enough for generalization?
	\item
	\textit{RQ4: Contributing Neurons.}
	Do detection models have neurons that are universally contributing across seen and unseen datasets?
\end{itemize}

Through RQ1, we would like to systematically assess the extent to which current deepfake image detectors can generalize in few-shot and (more importantly) zero-shot settings.
In cases where detectors do not generalize, through RQ2, we would like to know whether detection models have learned unwanted properties that interfere with their abilities to characterize deepfakes.
Through RQ3, we will apply AI interpretability techniques to assess the focus of detection models, i.e., whether they are extracting features that are discriminative enough for generalization.
Finally, through RQ4, we assess deepfake image detection models at a more granular level: through causality-based analysis, can we identify neurons that are contributing to the detection of both seen and unseen datasets?
The existence of such universally contributing neurons supports the prospect that detection models can focus on generalizable features.

\subsection{Dataset and Detector Selection}
\label{sec:selection}

In order to answer our above RQs, we require a representative selection of deepfake datasets and detectors, as in the following.

\textbf{Datasets.}
For our study, we would like to assess the generalizability across deepfakes generated by all of the state-of-the-art methods summarised in Table~\ref{tab:dfmethod}.
Thus, we selected the following datasets, which collectively cover all of the deepfake generation methods highlighted, and are frequently used to evaluate detectors in the literature.

\begin{itemize}
	\item \texttt{FaceForensics++}~\cite{rossler2019faceforensics++,faceforensic2019}.
	This benchmark consists of multiple sub-datasets, from which sets of 1000 deepfake videos were generated using each of the following synthesis methods: \texttt{FaceSwap} (\texttt{FS}), \texttt{DeepFake} (\texttt{DF}), 
	\texttt{Face2Face} (\texttt{F2F}), and \texttt{NeuralTextures} (\texttt{NT}).
	The videos are provided in multiple levels of quality, of which we select the high-quality ones (`C23') for our study.
	The dataset has been extensively used in evaluations, and has been cited over 2,200 times in the literature.
	
	\item \texttt{DeepFakeDetection (DFD)}~\cite{deepfakedetection2019}.
	Google AI released this dataset based on an improved version of the \texttt{FaceSwap-GAN} deepfake generation method.
	The authors invited 28 actors to record 363 real videos, creating 3068 fake ones using their method.
	The dataset has been incorporated into the aforementioned \texttt{FaceForensics++} benchmark, indicating the community's recognition of it as credible for evaluation.
	
	\item \texttt{CELEB}~\cite{celeb2019}.
	The authors proposed an improved version of the \texttt{FaceSwap-GAN} generation method and released two datasets resulting from it (\texttt{CELEB-DF-V1}~\cite{li2020celeb} and \texttt{CELEB-DF-V2}~\cite{li2019celeb}), which respectively contain 408 and 590 real videos as well as 795 and 5639 corresponding deepfakes. We utilize the latter in our study given the larger number of high-quality deepfake videos (as well as their involved frame images).
	\texttt{CELEB} is widely accepted for evaluation in the community, and has been cited over 1,200 times in the literature.
\end{itemize}

\textbf{Detectors.}
For our study, we selected state-of-the-art deep learning detectors (in Table~\ref{tab:detector}, to be specific).
Note that we do not select any of visual feature-based detectors, because they already have practical limitations before even considering generalization, as their requirement for specific features means that they fail to detect deepfakes reliably in newer datasets such as \texttt{CELEB}~\cite{li2020celeb}.
That is, the selected deep learning detectors are most practical and important for assessing their potential for generalization.

\begin{itemize}
	\item
	\texttt{MesoNet}, \texttt{MesoInception}~\cite{chattopadhyay2019neural}.
	Both detectors are based on CNN architectures, where \texttt{MesoNet} contains four convolutional and two dense layers and \texttt{MesoInception} contains two inception~\cite{szegedy2015going},
	two convolutional, and two dense layers.
	
	\item
	\texttt{ShallowNet}~\cite{tariq2019gan}.
	Among the three CNN-based detectors proposed by the authors, we select the \texttt{ShallowNet-V3} as our baseline detector since it reports the best performance.
	This detector consists of eight convolutional and two dense layers.
	
	\item
	\texttt{XceptionNet}~\cite{rossler2019faceforensics++}.
	This CNN-based detector consists of two convolutional and 13 separable convolutional~\cite{chollet2017xception} layers, as well as one dense layer.
	
	\item
	\texttt{EfficientNet}~\cite{tan2019efficientnet}.
	There exist several versions of this network, and we adopt \texttt{EfficientNet-B0} as our baseline model since it has been actually applied in practice~\cite{dfdetect2022}.
	This detector is based on a CNN architecture with
	two convolutional and seven MBConv~\cite{tan2019mnasnet} layers, followed by a dense layer.
\end{itemize}

\textbf{Model Augmentation Strategies.}
Recall our reviews in Section~\ref{sec:generalizability} that a number of works have investigated the use of model augmentation strategies to facilitate few-shot and (tentatively) zero-shot generalization.
Following a review of these works, we selected two model augmentation strategies that: (1)~can be applied to our selected deepfake image detectors; and (2)~are associated with the most promising results.
Based on these criteria, we selected transfer learning and merge learning, which achieved the most reasonable results across the works (in \texttt{OneRuleAll}~\cite{tariq2021one}, to be specific).
To expand, transfer learning is a typical model augmentation strategy that has been applied across several domains~\cite{zhuang2020comprehensive} for generalization.
One way to apply transfer learning is by fine-tuning.
Here, the weights of a pre-trained model (e.g., trained on one of our selected datasets) are frozen except for the last few layers, which are modified by extending knowledge from new datasets.
Merge learning, in contrast, is a more extreme strategy for generalization since it fine-tunes all the layers of a detection model using one or more additional datasets.

\subsection{Implementation}
\label{sec:impdetector}

\begin{table*}[t]
	\caption{Implemented deepfake image detectors.}
	\begin{center}
		\tiny
		\renewcommand\arraystretch{1.05}
		\resizebox{0.9\linewidth}{!}
		{
			\begin{tabular}{|c|c|cc|c|c|c|}
				\hline
				\multirow{2}{*}{Detector} & \multirow{2}{*}{Architecture}                          
				& \multicolumn{2}{c|}{Training   Dataset}                      
				& \multirow{2}{*}{Fine-tuning} 
				& \multirow{2}{*}{\begin{tabular}[c]{@{}c@{}}Original\\      
						Accuracy (\%)\end{tabular}} 
				& \multirow{2}{*}{\begin{tabular}[c]{@{}c@{}}Average\\      
						Accuracy (\%)\end{tabular}} \\ \cline{3-4}
				&  
				& \multicolumn{1}{c|}{Original}       
				& Augmented              
				&                              
				&                               
				&                     
				\\ \hline
				\texttt{XceptionNet1}     
				& \multirow{6}{*}{\begin{tabular}[c]{@{}c@{}}2-Conv\\      
						12-SeparableConv\\      \textbf{1-SeparableConv}\\      
						\textbf{1-Dense}\end{tabular}} 
				& \multicolumn{1}{c|}{\texttt{CELEB}} 
				& ——                     
				& ——                           
				& 99.10                                                                             
				& \multirow{6}{*}{95.93}   \\ \cline{1-1} \cline{3-6}
				\texttt{XceptionNet2} 
				&   
				& \multicolumn{1}{c|}{\texttt{CELEB}} 
				& \texttt{CELEB, FS, NT} 
				& Transfer
				& 98.45 
				&  \\ \cline{1-1} \cline{3-6}
				\texttt{XceptionNet3} 
				&   
				& \multicolumn{1}{c|}{\texttt{CELEB}} 
				& \texttt{CELEB, FS, NT} 
				& Merge  
				& 93.60 
				&  \\ \cline{1-1} \cline{3-6}
				\texttt{XceptionNet4} 
				&   
				& \multicolumn{1}{c|}{\texttt{F2F}}   
				& —— 
				& ——   
				& 96.15 
				&  \\ \cline{1-1} \cline{3-6}
				\texttt{XceptionNet5} 
				&   
				& \multicolumn{1}{c|}{\texttt{F2F}}   
				& \texttt{F2F, FS, NT}   
				& Transfer
				& 93.35 
				&  \\ \cline{1-1} \cline{3-6}
				\texttt{XceptionNet6} 
				&   
				& \multicolumn{1}{c|}{\texttt{F2F}}   
				& \texttt{F2F, FS, NT}   
				& Merge
				& 94.90 
				&  \\ \hline
				\texttt{MesoInception1}   
				& \multirow{6}{*}{\begin{tabular}[c]{@{}c@{}}2-Inception\\  
						2-Conv\\  \textbf{2-Dense}\end{tabular}} 
				& \multicolumn{1}{c|}{\texttt{CELEB}} 
				& —— 
				& ——   
				& 97.70 
				& \multirow{6}{*}{95.23}   \\ \cline{1-1} \cline{3-6}
				\texttt{MesoInception2}   
				&   
				& \multicolumn{1}{c|}{\texttt{CELEB}} 
				& \texttt{CELEB, FS, NT} 
				& Transfer
				& 96.50 
				&  \\ \cline{1-1} \cline{3-6}
				\texttt{MesoInception3}   
				&   
				& \multicolumn{1}{c|}{\texttt{CELEB}} 
				& \texttt{CELEB, FS, NT} 
				& Merge  
				& 89.05 
				&  \\ \cline{1-1} \cline{3-6}
				\texttt{MesoInception4}   
				&   
				& \multicolumn{1}{c|}{\texttt{F2F}}   
				& —— 
				& ——   
				& 97.60 
				&  \\ \cline{1-1} \cline{3-6}
				\texttt{MesoInception5}   
				&   
				& \multicolumn{1}{c|}{\texttt{F2F}}   
				& \texttt{F2F, FS, NT}   
				& Transfer
				& 96.35 
				&  \\ \cline{1-1} \cline{3-6}
				\texttt{MesoInception6}   
				&   
				& \multicolumn{1}{c|}{\texttt{F2F}}   
				& \texttt{F2F, FS, NT}   
				& Merge 
				& 94.15 
				&  \\ \hline
				\texttt{MesoNet1} 
				& \multirow{6}{*}{\begin{tabular}[c]{@{}c@{}}4-Conv\\  
						\textbf{2-Dense}\end{tabular}}
				& \multicolumn{1}{c|}{\texttt{CELEB}} 
				& —— 
				& ——   
				& 97.60 
				& \multirow{6}{*}{94.69}   \\ \cline{1-1} \cline{3-6}
				\texttt{MesoNet2} 
				&   
				& \multicolumn{1}{c|}{\texttt{CELEB}} 
				& \texttt{CELEB, FS, NT} 
				& Transfer
				& 96.85 
				&  \\ \cline{1-1} \cline{3-6}
				\texttt{MesoNet3} 
				&   
				& \multicolumn{1}{c|}{\texttt{CELEB}} 
				& \texttt{CELEB, FS, NT} 
				& Merge  
				& 85.50 
				&  \\ \cline{1-1} \cline{3-6}
				\texttt{MesoNet4} 
				&   
				& \multicolumn{1}{c|}{\texttt{F2F}}   
				& —— 
				& ——   
				& 97.35 
				&  \\ \cline{1-1} \cline{3-6}
				\texttt{MesoNet5} 
				&   
				& \multicolumn{1}{c|}{\texttt{F2F}}   
				& \texttt{F2F, FS, NT}   
				& Transfer
				& 96.20 
				&  \\ \cline{1-1} \cline{3-6}
				\texttt{MesoNet6} 
				&   
				& \multicolumn{1}{c|}{\texttt{F2F}}   
				& \texttt{F2F, FS, NT}   
				& Merge
				& 94.65 
				&  \\ \hline
				\texttt{EfficientNet1}
				& \multirow{6}{*}{\begin{tabular}[c]{@{}c@{}}1-Conv\\  
						7-MBConv \\  \textbf{1-Conv}\\  \textbf{1-Dense}\end{tabular}} 
				& \multicolumn{1}{c|}{\texttt{CELEB}} 
				& —— 
				& ——   
				& 88.45 
				& \multirow{6}{*}{81.35}   \\ \cline{1-1} \cline{3-6}
				\texttt{EfficientNet2}
				&   
				& \multicolumn{1}{c|}{\texttt{CELEB}} 
				& \texttt{CELEB, FS, NT} 
				& Transfer
				& 88.15 
				&  \\ \cline{1-1} \cline{3-6}
				\texttt{EfficientNet3}
				&   
				& \multicolumn{1}{c|}{\texttt{CELEB}} 
				& \texttt{CELEB, FS, NT} 
				& Merge   
				& 79.05 
				&  \\ \cline{1-1} \cline{3-6}
				\texttt{EfficientNet4}
				&   
				& \multicolumn{1}{c|}{\texttt{F2F}}   
				& —— 
				& ——   
				& 78.50
				&  \\ \cline{1-1} \cline{3-6}
				\texttt{EfficientNet5}
				&   
				& \multicolumn{1}{c|}{\texttt{F2F}}   
				& \texttt{F2F, FS, NT}   
				& Transfer
				& 76.80 
				&  \\ \cline{1-1} \cline{3-6}
				\texttt{EfficientNet6}
				&   
				& \multicolumn{1}{c|}{\texttt{F2F}}   
				& \texttt{F2F, FS, NT}   
				& Merge  
				& 77.15 
				&  \\ \hline
				\texttt{ShallowNet1}  
				& \multirow{6}{*}{\begin{tabular}[c]{@{}c@{}}8-Conv\\  
						\textbf{2-Dense}\end{tabular}}
				& \multicolumn{1}{c|}{\texttt{CELEB}} 
				& —— 
				& ——   
				& 85.40 
				& \multirow{6}{*}{77.70}   \\ \cline{1-1} \cline{3-6}
				\texttt{ShallowNet2}  
				&   
				& \multicolumn{1}{c|}{\texttt{CELEB}} 
				& \texttt{CELEB, FS, NT} 
				& Transfer
				& 71.15 
				&  \\ \cline{1-1} \cline{3-6}
				\texttt{ShallowNet3}  
				&   
				& \multicolumn{1}{c|}{\texttt{CELEB}} 
				& \texttt{CELEB, FS, NT} 
				& Merge  
				& 73.20 
				&  \\ \cline{1-1} \cline{3-6}
				\texttt{ShallowNet4}  
				&   
				& \multicolumn{1}{c|}{\texttt{F2F}}   
				& —— 
				& ——   
				& 88.85 
				&  \\ \cline{1-1} \cline{3-6}
				\texttt{ShallowNet5}  
				&   
				& \multicolumn{1}{c|}{\texttt{F2F}}   
				& \texttt{F2F, FS, NT}   
				& Transfer
				& 76.60 
				&  \\ \cline{1-1} \cline{3-6}
				\texttt{ShallowNet6}  
				&   
				& \multicolumn{1}{c|}{\texttt{F2F}}   
				& \texttt{F2F, FS, NT}   
				& Merge  
				& 71.00 
				&  \\ \hline
			\end{tabular}
		}\label{tab:implementation}
	\end{center}
\end{table*}

To facilitate our experiments, we trained 30 deepfake image detectors using the models and datasets identified in Section~\ref{sec:selection}.
In particular, for each of the five detectors, we trained a model on the \texttt{CELEB} dataset and another on the \texttt{F2F} data extracted from \texttt{FaceForensics++}, and then augmented the models using either transfer or merge learning based on additional samples from the training as well as unseen ones (\texttt{FS}, \texttt{NT}).
Note that \texttt{CELEB} and \texttt{F2F} are selected for model training, since they respectively belong to categories of face swapping and face reenactment, and are generated based on deep learning and computer graphics methods, which allows for fair and comprehensive evaluations (the selection of \texttt{FS} and \texttt{NT} for fine-tuning follows the same considerations).
Table~\ref{tab:implementation} summarizes the resulting detectors, and we present more details on their training in the following.

First, as a pre-processing step, we utilized a face recognition project~\cite{facerecognition2018} to extract 10 frame images from each video in our selected datasets, align and crop their facial regions, and save the cropped images in PNG format with a size of $256 \times 256$.

Next, to train the `original' deepfake image detectors such as \texttt{MesoNet1} and \texttt{MesoNet4}, i.e., without any model augmentation, we used a training set of 16,000 images from the (respective) original datasets.
To further train the fine-tuned detectors (such as \texttt{MesoNet2} and \texttt{MesoNet3}), the augmented training set consisted of 2,000 additional images from the original dataset as well as 2,000 from \texttt{FS} and \texttt{NT}.
When performing transfer learning, we fine-tuned the last two function layers of these original detectors, or other parts of architectures that contain significant features for classification (indicated in \textbf{bold} in Table~\ref{tab:implementation}).

Each detector in the table was trained based on an \texttt{Adam}~\cite{kingma2014adam} optimizer with $\beta1=0.9$ and $\beta2=0.999$.
The learning rate for \texttt{MesoNet} and \texttt{MesoInception} detectors was $10^{-3}$, whereas for the remaining detectors it was $5 \times 10^{-5}$.
Moreover, the original and augmented detectors were respectively trained for 100 and 50 epochs, and the best-performing parameters were adopted (instead of the last-epoch ones).
All training and evaluations were conducted on a server with an Intel Core i9-9900 3.1GHz CPU, 128GB memory, and two NVIDIA GeForce RTX 2080Ti graphics cards.

Table~\ref{tab:implementation} also shows the accuracy of each implemented detector.
The `original accuracy' was evaluated based on testing images extracted from \texttt{CELEB} or \texttt{F2F} respectively.
For our study, each testing dataset is composed of 2,000 images that are fully independent of the training ones.
We remark that each of our training and testing datasets consists of 50\% real and 50\% deepfake images.
As seen in the table, most deepfake image detectors have promising accuracy, although a number of the fine-tuned ones, e.g., \texttt{ShallowNet2}, show comparatively poor results (below 80\%).
In these cases, fine-tuning inevitably impacted the original performance of the models, since learning new knowledge unavoidably weakens their attention to the features in original datasets.

\section{Empirical Study}
\label{study}

In this section, we present the experiments designed to answer our RQs (as in Section~\ref{sec:research_questions}), analyze the results, and summarize our findings.

\begin{table*}[t]
	\caption{Generalizability of deepfake image detectors to unseen datasets.}
	\begin{center}
		\tiny
		\renewcommand\arraystretch{1}
		\resizebox{0.85\linewidth}{!}
		{
			\begin{tabular}{|c|c|c|cccccc|}
				\hline
				\multirow{3}{*}{Rank}
				& \multirow{3}{*}{Detector}
				& \multirow{3}{*}{\begin{tabular}[c]{@{}c@{}}
						Original\\Accuracy (\%)\end{tabular}}
				& \multicolumn{6}{c|}{Unseen Dataset} 
				\\ \cline{4-9} 
				&
				&& \multicolumn{5}{c|}{Accuracy (\%)}& \multirow{2}{*}
				{\begin{tabular}[c]{@{}c@{}}Degradation\\(\% points)\end{tabular}} 
				\\ \cline{4-8}
				&
				&& \multicolumn{1}{c|}{\texttt{FS}}& \multicolumn{1}{c|}
				{\texttt{NT}}& \multicolumn{1}{c|}{\texttt{DF}}
				& \multicolumn{1}{c|}{\texttt{DFD}}
				& \multicolumn{1}{c|}{Average}
				& \\ \hline
				1                     
				& \texttt{ShallowNet6}      
				& 71.00                                                                                  
				& \multicolumn{1}{c|}{79.50}       
				& \multicolumn{1}{c|}{75.60}       
				& \multicolumn{1}{c|}{56.20}       
				& \multicolumn{1}{c|}{52.10}        
				& \multicolumn{1}{c|}{65.85}   
				& 5.15 \\ \hline
				2                     
				& \texttt{ShallowNet2}      
				& 71.15 
				& \multicolumn{1}{c|}{75.60}       
				& \multicolumn{1}{c|}{70.85}       
				& \multicolumn{1}{c|}{53.60}       
				& \multicolumn{1}{c|}{51.75}        
				& \multicolumn{1}{c|}{62.95}   
				& 8.20 \\ \hline
				3                     
				& \texttt{ShallowNet3}      
				& 73.20 
				& \multicolumn{1}{c|}{75.60}       
				& \multicolumn{1}{c|}{71.40}       
				& \multicolumn{1}{c|}{53.95}       
				& \multicolumn{1}{c|}{55.60}        
				& \multicolumn{1}{c|}{64.14}   
				& 9.06 \\ \hline
				4                     
				& \texttt{ShallowNet5}      
				& 76.60 
				& \multicolumn{1}{c|}{78.55}       
				& \multicolumn{1}{c|}{77.75}       
				& \multicolumn{1}{c|}{57.20}       
				& \multicolumn{1}{c|}{55.85}        
				& \multicolumn{1}{c|}{67.34}   
				& 9.26 \\ \hline
				5                     
				& \texttt{EfficientNet6}    
				& 77.15 
				& \multicolumn{1}{c|}{75.05}       
				& \multicolumn{1}{c|}{76.15}       
				& \multicolumn{1}{c|}{55.30}       
				& \multicolumn{1}{c|}{46.15}        
				& \multicolumn{1}{c|}{63.16}   
				& 13.99\\ \hline
				6                     
				& \texttt{MesoNet3}         
				& 85.50 
				& \multicolumn{1}{c|}{75.50}       
				& \multicolumn{1}{c|}{76.10}       
				& \multicolumn{1}{c|}{65.95}       
				& \multicolumn{1}{c|}{61.45}        
				& \multicolumn{1}{c|}{69.75}   
				& 15.75\\ \hline
				7                     
				& \texttt{EfficientNet3}    
				& 79.05 
				& \multicolumn{1}{c|}{67.70}       
				& \multicolumn{1}{c|}{66.35}       
				& \multicolumn{1}{c|}{63.05}       
				& \multicolumn{1}{c|}{50.35}        
				& \multicolumn{1}{c|}{61.86}   
				& 17.19\\ \hline
				8                     
				& \texttt{MesoInception3}   
				& 89.05 
				& \multicolumn{1}{c|}{76.25}       
				& \multicolumn{1}{c|}{71.05}       
				& \multicolumn{1}{c|}{59.50}       
				& \multicolumn{1}{c|}{56.95}        
				& \multicolumn{1}{c|}{65.94}   
				& 23.11\\ \hline
				9                     
				& \texttt{EfficientNet5}    
				& 76.80 
				& \multicolumn{1}{c|}{51.25}       
				& \multicolumn{1}{c|}{54.30}       
				& \multicolumn{1}{c|}{56.30}       
				& \multicolumn{1}{c|}{49.40}        
				& \multicolumn{1}{c|}{52.81}   
				& 23.99\\ \hline
				10                    
				& \texttt{XceptionNet3}     
				& 93.60 
				& \multicolumn{1}{c|}{80.75}       
				& \multicolumn{1}{c|}{76.35}       
				& \multicolumn{1}{c|}{63.90}       
				& \multicolumn{1}{c|}{55.10}        
				& \multicolumn{1}{c|}{69.03}   
				& 24.58\\ \hline
				11                    
				& \texttt{XceptionNet6}     
				& 94.90 
				& \multicolumn{1}{c|}{83.95}       
				& \multicolumn{1}{c|}{83.40}       
				& \multicolumn{1}{c|}{58.50}       
				& \multicolumn{1}{c|}{53.25}        
				& \multicolumn{1}{c|}{69.78}   
				& 25.13\\ \hline
				12                    
				& \texttt{MesoNet6}         
				& 94.65 
				& \multicolumn{1}{c|}{77.80}       
				& \multicolumn{1}{c|}{76.90}       
				& \multicolumn{1}{c|}{60.95}       
				& \multicolumn{1}{c|}{57.95}        
				& \multicolumn{1}{c|}{68.40}   
				& 26.25\\ \hline
				13                    
				& \texttt{EfficientNet4}    
				& 78.50 
				& \multicolumn{1}{c|}{49.10}       
				& \multicolumn{1}{c|}{53.90}       
				& \multicolumn{1}{c|}{55.85}       
				& \multicolumn{1}{c|}{48.80}        
				& \multicolumn{1}{c|}{51.91}   
				& 26.59\\ \hline
				14                    
				& \texttt{MesoInception6}   
				& 94.15 
				& \multicolumn{1}{c|}{82.05}       
				& \multicolumn{1}{c|}{78.25}       
				& \multicolumn{1}{c|}{56.45}       
				& \multicolumn{1}{c|}{52.70}        
				& \multicolumn{1}{c|}{67.36}   
				& 26.79\\ \hline
				15                    
				& \texttt{ShallowNet1}      
				& 85.40 
				& \multicolumn{1}{c|}{58.05}       
				& \multicolumn{1}{c|}{50.20}       
				& \multicolumn{1}{c|}{54.90}       
				& \multicolumn{1}{c|}{49.60}        
				& \multicolumn{1}{c|}{53.19}   
				& 32.21\\ \hline
				16                    
				& \texttt{ShallowNet4}      
				& 88.85 
				& \multicolumn{1}{c|}{53.50}       
				& \multicolumn{1}{c|}{53.35}       
				& \multicolumn{1}{c|}{58.45}       
				& \multicolumn{1}{c|}{55.15}        
				& \multicolumn{1}{c|}{55.11}   
				& 33.74\\ \hline
				17                    
				& \texttt{EfficientNet2}    
				& 88.15 
				& \multicolumn{1}{c|}{50.30}       
				& \multicolumn{1}{c|}{50.40}       
				& \multicolumn{1}{c|}{58.25}       
				& \multicolumn{1}{c|}{52.60}        
				& \multicolumn{1}{c|}{52.89}   
				& 35.26\\ \hline
				18                    
				& \texttt{EfficientNet1}    
				& 88.45 
				& \multicolumn{1}{c|}{51.95}       
				& \multicolumn{1}{c|}{49.80}       
				& \multicolumn{1}{c|}{58.70}       
				& \multicolumn{1}{c|}{51.95}        
				& \multicolumn{1}{c|}{53.10}   
				& 35.35\\ \hline
				19                    
				& \texttt{XceptionNet5}     
				& 93.35 
				& \multicolumn{1}{c|}{60.00}       
				& \multicolumn{1}{c|}{62.55}       
				& \multicolumn{1}{c|}{56.70}       
				& \multicolumn{1}{c|}{47.25}        
				& \multicolumn{1}{c|}{56.63}   
				& 36.73\\ \hline
				20                    
				& \texttt{MesoNet5}         
				& 96.20 
				& \multicolumn{1}{c|}{64.55}       
				& \multicolumn{1}{c|}{62.40}       
				& \multicolumn{1}{c|}{56.75}       
				& \multicolumn{1}{c|}{53.55}        
				& \multicolumn{1}{c|}{59.31}   
				& 36.89\\ \hline
				21                    
				& \texttt{MesoInception5}   
				& 96.35 
				& \multicolumn{1}{c|}{57.45}       
				& \multicolumn{1}{c|}{60.60}       
				& \multicolumn{1}{c|}{60.50}       
				& \multicolumn{1}{c|}{55.50}        
				& \multicolumn{1}{c|}{58.51}   
				& 37.84\\ \hline
				22                    
				& \texttt{MesoNet2}         
				& 96.85 
				& \multicolumn{1}{c|}{57.55}       
				& \multicolumn{1}{c|}{58.35}       
				& \multicolumn{1}{c|}{60.80}       
				& \multicolumn{1}{c|}{55.65}        
				& \multicolumn{1}{c|}{58.09}   
				& 38.76\\ \hline
				23                    
				& \texttt{MesoInception2}   
				& 96.50 
				& \multicolumn{1}{c|}{54.65}       
				& \multicolumn{1}{c|}{55.25}       
				& \multicolumn{1}{c|}{56.85}       
				& \multicolumn{1}{c|}{51.60}        
				& \multicolumn{1}{c|}{54.59}   
				& 41.91\\ \hline
				24                    
				& \texttt{MesoNet1}         
				& 97.60 
				& \multicolumn{1}{c|}{51.15}       
				& \multicolumn{1}{c|}{52.30}       
				& \multicolumn{1}{c|}{61.80}       
				& \multicolumn{1}{c|}{55.75}        
				& \multicolumn{1}{c|}{55.25}   
				& 42.35\\ \hline
				25                    
				& \texttt{XceptionNet4}     
				& 96.15 
				& \multicolumn{1}{c|}{48.80}       
				& \multicolumn{1}{c|}{55.75}       
				& \multicolumn{1}{c|}{57.30}       
				& \multicolumn{1}{c|}{50.80}        
				& \multicolumn{1}{c|}{53.16}   
				& 42.99\\ \hline
				26                    
				& \texttt{MesoNet4}         
				& 97.35 
				& \multicolumn{1}{c|}{54.55}       
				& \multicolumn{1}{c|}{52.20}       
				& \multicolumn{1}{c|}{56.00}       
				& \multicolumn{1}{c|}{54.35}        
				& \multicolumn{1}{c|}{54.28}   
				& 43.08\\ \hline
				27                    
				& \texttt{MesoInception4}   
				& 97.60 
				& \multicolumn{1}{c|}{50.00}       
				& \multicolumn{1}{c|}{53.35}       
				& \multicolumn{1}{c|}{59.10}       
				& \multicolumn{1}{c|}{53.95}    
				& \multicolumn{1}{c|}{54.10}   
				& 43.50\\ \hline
				28
				& \texttt{MesoInception1}   
				& 97.70 
				& \multicolumn{1}{c|}{52.80}   
				& \multicolumn{1}{c|}{52.10}   
				& \multicolumn{1}{c|}{57.10}
				& \multicolumn{1}{c|}{51.90} 
				& \multicolumn{1}{c|}{53.48}   
				& 44.23\\ \hline
				29  
				& \texttt{XceptionNet2} 
				& 98.45 
				& \multicolumn{1}{c|}{55.25}
				& \multicolumn{1}{c|}{51.15}
				& \multicolumn{1}{c|}{56.80}
				& \multicolumn{1}{c|}{53.45} 
				& \multicolumn{1}{c|}{54.16}   
				& 44.29\\ \hline
				30  
				& \texttt{XceptionNet1} 
				& 99.10 
				& \multicolumn{1}{c|}{54.60}
				& \multicolumn{1}{c|}{50.95}
				& \multicolumn{1}{c|}{57.00}
				& \multicolumn{1}{c|}{54.80} 
				& \multicolumn{1}{c|}{54.34}   
				& 44.76\\ \hline
			\end{tabular}
		}\label{tab:result1}
	\end{center}
\end{table*}

\subsection{Generalizability of Deepfake Image Detectors (RQ1)}
\label{RQ1}

\textbf{Experiments.}
Recall our goal to determine the extent to which detectors generalize across deepfake datasets other than those they trained on.
To achieve this, we took each of the 30 implemented detectors described in Section~\ref{sec:impdetector} and evaluated their accuracy on the unseen datasets described in Section~\ref{sec:selection}, i.e., \texttt{FS}, \texttt{NT}, \texttt{DF}, and \texttt{DFD}.
Note that although a number of samples from \texttt{FS} and \texttt{NT} have participated in the fine-tuning of partial augmented detectors (e.g., \texttt{MesoNet2} and \texttt{MesoNet3}), we still consider the independent testing images from the two datasets as unseen ones, since this is a general `few-shot' setting.

\emph{Measure of generalizability.}
We present our evaluation results in Table~\ref{tab:result1}.
As the rightmost columns in the table, we measure the generalizability of a detector by computing its performance degradation, i.e., the difference between its original accuracy (i.e., on the original testing dataset) and the average accuracy it achieved across different unseen datasets.
Note that for the performance degradation, lower values indicate better generalizability, whereas higher values indicate that the respective detectors generalize poorly to the given unseen datasets.

\textbf{Result Analysis.}
First, in Table~\ref{tab:result1}, we observe that most of these detectors (at ranks \#8--30) generalize poorly, with all but the top seven of our detectors seeing a performance degradation of over 20 percentage points (and as many as 44.76 percentage points for \texttt{XceptionNet1}).
Upon inspecting the accuracy of detectors on specific unseen datasets, we observe that most of them are below $70\%$, with multiple results close to $50\%$ which can be considered as random guesses.
For example, the rank \#30 \texttt{XceptionNet1} achieves only 54.60\% accuracy on \texttt{FS}, 50.95\% on \texttt{NT}, 57.00\% on \texttt{DF} and 54.80\% on \texttt{DFD}.
Furthermore, although the detectors at ranks \#1--7 achieve passable performance degradation (below 20 percentage points), their accuracies on the fully unseen datasets remain below $70\%$.
For example, \texttt{MesoInception3} achieves accuracies of 59.50\% and 56.95\% on \texttt{DF} and \texttt{DFD}.
Such results confirm the inability of these detectors to generalize in zero-shot settings.

Second, to further observe specific results, the accuracy of several detectors is over $70\%$ on the unseen \texttt{FS} and \texttt{NT} datasets.
For example, on \texttt{FS} and \texttt{NT}, \texttt{ShallowNet6} achieves accuracies of 79.50\% and 75.60\%, and \texttt{MesoInception3} achieves accuracies of 76.25\% and 71.05\%.
Recall that these detectors were augmented using merge learning, in which a total of 1000 deepfake images from \texttt{FS} and \texttt{NT} were used for fine-tuning.
This suggests these detectors can generalize to particular unseen datasets with the help of a few unseen deepfakes.

In general, these results are consistent with those previously published results on few-shot and zero-shot generalizability (Section~\ref{sec:generalizability}).\\

\begin{tabular}{lp{2.2in}}
	\textbf{\underline{Finding 1}:} & Existing deepfake image detectors are not generalizable in zero-shot settings, and the performance degradation can be mitigated in few-shot settings.
\end{tabular}

\subsection{Learning Unwanted Properties (RQ2)}
\label{RQ2}

Given that detectors do not generalize except in some few-shot settings (Section~\ref{RQ1}), it is intuitive that they may have learned certain unwanted properties rather than those intrinsic ones in deepfakes.
For example, if detectors are learning properties that pertain to specific synthesis methods or human identities, this might interfere with learning the essential difference between real and fake samples.
We explore this intuition in RQ2.

\begin{table*}[t]
	\caption{Accuracy of detectors on an unseen dataset synthesized by the same method as the training set.}
	\begin{center}
		\renewcommand\arraystretch{1}
		\resizebox{0.9\linewidth}{!}
		{
			\begin{tabular}{|c|c|c|cccccc|cc|c|}
				\hline
				\multirow{3}{*}{Rank} 
				& \multirow{3}{*}{Detector} 
				& \multirow{3}{*}{\begin{tabular}[c]{@{}c@{}}
						Original\\Accuracy\\ (\%)\end{tabular}} 
				& \multicolumn{6}{c|}{Unseen Dataset} 
				& \multicolumn{2}{c|}{Specialized Dataset} 
				& \multirow{3}{*}{\begin{tabular}[c]{@{}c@{}}
						Degradation\\Difference\\(\% points)\end{tabular}} 
				\\ \cline{4-11}
				& 
				&
				& \multicolumn{5}{c|}{Accuracy (\%)}
				& \multirow{2}{*}{\begin{tabular}[c]{@{}c@{}}
						Degradation\\(\% points)\end{tabular}} 
				& \multicolumn{1}{c|}{Accuracy (\%)} 
				& \multirow{2}{*}{\begin{tabular}[c]{@{}c@{}}
						Degradation\\(\% points)\end{tabular}} 
				&\\ \cline{4-8} \cline{10-10}
				& 
				&
				& \multicolumn{1}{c|}{\texttt{FS}} 
				& \multicolumn{1}{c|}{\texttt{NT}} 
				& \multicolumn{1}{c|}{\texttt{DF}} 
				& \multicolumn{1}{c|}{\texttt{DFD}} 
				& \multicolumn{1}{c|}{Average} 
				& 
				& \multicolumn{1}{c|}{\texttt{CELEB-M}} 
				& 
				&\\ \hline
				1 
				& \texttt{MesoNet1} 
				& 97.60
				& \multicolumn{1}{c|}{51.15} 
				& \multicolumn{1}{c|}{52.30} 
				& \multicolumn{1}{c|}{61.80} 
				& \multicolumn{1}{c|}{55.75}
				& \multicolumn{1}{c|}{55.25} 
				& 42.35 
				& \multicolumn{1}{c|}{80.20}
				& 17.40 
				& 24.95 \\ \hline
				2 
				& \texttt{XceptionNet1} 
				& 99.10
				& \multicolumn{1}{c|}{54.60} 
				& \multicolumn{1}{c|}{50.95} 
				& \multicolumn{1}{c|}{57.00} 
				& \multicolumn{1}{c|}{54.80}
				& \multicolumn{1}{c|}{54.34} 
				& 44.76 
				& \multicolumn{1}{c|}{76.05}
				& 23.05 
				& 21.71 \\ \hline
				3 
				& \texttt{XceptionNet2} 
				& 98.45
				& \multicolumn{1}{c|}{55.25} 
				& \multicolumn{1}{c|}{51.15} 
				& \multicolumn{1}{c|}{56.80} 
				& \multicolumn{1}{c|}{53.45}
				& \multicolumn{1}{c|}{54.16} 
				& 44.29 
				& \multicolumn{1}{c|}{75.15}
				& 23.30 
				& 20.99 \\ \hline
				4 
				& \texttt{MesoInception1} 
				& 97.70
				& \multicolumn{1}{c|}{52.80} 
				& \multicolumn{1}{c|}{52.10} 
				& \multicolumn{1}{c|}{57.10} 
				& \multicolumn{1}{c|}{51.90}
				& \multicolumn{1}{c|}{53.48} 
				& 44.23 
				& \multicolumn{1}{c|}{74.25}
				& 23.45 
				& 20.78 \\ \hline
				5 
				& \texttt{MesoNet2} 
				& 96.85
				& \multicolumn{1}{c|}{57.55} 
				& \multicolumn{1}{c|}{58.35} 
				& \multicolumn{1}{c|}{60.80} 
				& \multicolumn{1}{c|}{55.65}
				& \multicolumn{1}{c|}{58.09} 
				& 38.76 
				& \multicolumn{1}{c|}{77.00}
				& 19.85 
				& 18.91 \\ \hline
				6 
				& \texttt{MesoInception2} 
				& 96.50
				& \multicolumn{1}{c|}{54.65} 
				& \multicolumn{1}{c|}{55.25} 
				& \multicolumn{1}{c|}{56.85} 
				& \multicolumn{1}{c|}{51.60}
				& \multicolumn{1}{c|}{54.59} 
				& 41.91 
				& \multicolumn{1}{c|}{71.75}
				& 24.75 
				& 17.16 \\ \hline
				7 
				& \texttt{ShallowNet1}
				& 85.40
				& \multicolumn{1}{c|}{58.05} 
				& \multicolumn{1}{c|}{50.20} 
				& \multicolumn{1}{c|}{54.90} 
				& \multicolumn{1}{c|}{49.60}
				& \multicolumn{1}{c|}{53.19} 
				& 32.21 
				& \multicolumn{1}{c|}{62.85} 
				& 22.55 
				& 9.66 \\ \hline
				8 
				& \texttt{EfficientNet2}
				& 88.15
				& \multicolumn{1}{c|}{50.30} 
				& \multicolumn{1}{c|}{50.40} 
				& \multicolumn{1}{c|}{58.25} 
				& \multicolumn{1}{c|}{52.60}
				& \multicolumn{1}{c|}{52.89} 
				& 35.26 
				& \multicolumn{1}{c|}{60.70} 
				& 27.45 
				& 7.81 \\ \hline
				9
				& \texttt{EfficientNet1}
				& 88.45
				& \multicolumn{1}{c|}{51.95} 
				& \multicolumn{1}{c|}{49.80} 
				& \multicolumn{1}{c|}{58.70} 
				& \multicolumn{1}{c|}{51.95}
				& \multicolumn{1}{c|}{53.10} 
				& 35.35 
				& \multicolumn{1}{c|}{60.55} 
				& 27.90 
				& 7.45 \\ \hline
			\end{tabular}
		}\label{tab:result2-1}
	\end{center}
\end{table*}

\begin{table*}[t]
	\caption{Accuracy of detectors on unseen datasets containing the same human identities as the training set.}
	\begin{center}
        \Large
		\renewcommand\arraystretch{1.05}
		\resizebox{0.9\linewidth}{!}
		{
			\begin{tabular}{|c|c|c|cccccc|cccc|c|}
				\hline
				\multirow{3}{*}{Rank} & \multirow{3}{*}{Detector} 
				& \multirow{3}{*}{\begin{tabular}[c]{@{}c@{}}Original\\ 
						Accuracy\\(\%)\end{tabular}} & \multicolumn{6}{c|}{Unseen Dataset} 
				& \multicolumn{4}{c|}{Specialized Dataset}           
				&\multirow{3}{*}{\begin{tabular}[c]{@{}c@{}}Degradation\\
						Difference\\(\% points)\end{tabular}}\\\cline{4-13}
				&
				&
				&\multicolumn{5}{c|}{Accuracy (\%)}
				&\multirow{2}{*}{\begin{tabular}[c]{@{}c@{}}
						Degradation\\(\% points)\end{tabular}}
				&\multicolumn{3}{c|}{Accuracy (\%)}
				&\multirow{2}{*}{\begin{tabular}[c]{@{}c@{}}
						Degradation\\(\% points)\end{tabular}}
				&\\\cline{4-8}\cline{10-12}
				&
				&
				&\multicolumn{1}{c|}{\texttt{FS}}
				&\multicolumn{1}{c|}{\texttt{NT}}
				&\multicolumn{1}{c|}{\texttt{DF}}
				&\multicolumn{1}{c|}{\texttt{DFD}}
				&\multicolumn{1}{c|}{Average}
				&
				&\multicolumn{1}{c|}{\texttt{FS-I}}
				&\multicolumn{1}{c|}{\texttt{NT-I}}
				&\multicolumn{1}{c|}{Average}
				&
				&\\\hline
				1 
				& \texttt{ShallowNet4}  
				& 88.85  
				& \multicolumn{1}{c|}{53.50}   
				& \multicolumn{1}{c|}{53.35}   
				& \multicolumn{1}{c|}{58.45}   
				& \multicolumn{1}{c|}{55.15}
				& \multicolumn{1}{c|}{55.11}   
				& 33.74 
				& \multicolumn{1}{c|}{66.95}  
				& \multicolumn{1}{c|}{61.15}  
				& \multicolumn{1}{c|}{64.05} 
				& \multicolumn{1}{c|}{24.80} 
				& 8.94  \\ \hline
				2 
				& \texttt{XceptionNet5} 
				& 93.35  
				& \multicolumn{1}{c|}{60.00}   
				& \multicolumn{1}{c|}{62.55}   
				& \multicolumn{1}{c|}{56.70}   
				& \multicolumn{1}{c|}{47.25}
				& \multicolumn{1}{c|}{56.63}   
				& 36.73 
				& \multicolumn{1}{c|}{64.40}  
				& \multicolumn{1}{c|}{61.00}  
				& \multicolumn{1}{c|}{62.70} 
				& \multicolumn{1}{c|}{30.65} 
				& 6.08  \\ \hline
				3 
				& \texttt{XceptionNet4} 
				& 96.15  
				& \multicolumn{1}{c|}{48.80}   
				& \multicolumn{1}{c|}{55.75}   
				& \multicolumn{1}{c|}{57.30}   
				& \multicolumn{1}{c|}{50.80}
				& \multicolumn{1}{c|}{53.16}   
				& 42.99 
				& \multicolumn{1}{c|}{57.80}  
				& \multicolumn{1}{c|}{55.65}  
				& \multicolumn{1}{c|}{56.73} 
				& \multicolumn{1}{c|}{39.43} 
				& 3.56  \\ \hline
				4 
				& \texttt{MesoInception4}   
				& 97.60  
				& \multicolumn{1}{c|}{50.00}   
				& \multicolumn{1}{c|}{53.35}   
				& \multicolumn{1}{c|}{59.10}   
				& \multicolumn{1}{c|}{53.95}
				& \multicolumn{1}{c|}{54.10}   
				& 43.50 
				& \multicolumn{1}{c|}{53.90}  
				& \multicolumn{1}{c|}{53.85}  
				& \multicolumn{1}{c|}{53.88} 
				& \multicolumn{1}{c|}{43.73} 
				& -0.22 \\ \hline
				5 
				& \texttt{MesoNet4} 
				& 97.35  
				& \multicolumn{1}{c|}{54.55}   
				& \multicolumn{1}{c|}{52.20}   
				& \multicolumn{1}{c|}{56.00}   
				& \multicolumn{1}{c|}{54.35}
				& \multicolumn{1}{c|}{54.28}   
				& 43.08 
				& \multicolumn{1}{c|}{55.25}  
				& \multicolumn{1}{c|}{52.35}  
				& \multicolumn{1}{c|}{53.80} 
				& \multicolumn{1}{c|}{43.55} 
				& -0.48 \\ \hline
				6 
				& \texttt{MesoNet5} 
				& 96.20  
				& \multicolumn{1}{c|}{64.55}   
				& \multicolumn{1}{c|}{62.40}   
				& \multicolumn{1}{c|}{56.75}   
				& \multicolumn{1}{c|}{53.55}
				& \multicolumn{1}{c|}{59.31}   
				& 36.89 
				& \multicolumn{1}{c|}{59.80}  
				& \multicolumn{1}{c|}{57.25}  
				& \multicolumn{1}{c|}{58.53} 
				& \multicolumn{1}{c|}{37.68} 
				& -0.79 \\ \hline
				7 
				& \texttt{MesoInception5}   
				& 96.35  
				& \multicolumn{1}{c|}{57.45}   
				& \multicolumn{1}{c|}{60.60}   
				& \multicolumn{1}{c|}{60.50}   
				& \multicolumn{1}{c|}{55.50}
				& \multicolumn{1}{c|}{58.51}   
				& 37.84 
				& \multicolumn{1}{c|}{54.50}  
				& \multicolumn{1}{c|}{56.15}  
				& \multicolumn{1}{c|}{55.33} 
				& \multicolumn{1}{c|}{41.03} 
				& -3.19 \\ \hline
			\end{tabular}
		}\label{tab:result2-2}
	\end{center}
\end{table*}

\textbf{Experiments.}
First, we selected the detectors trained on \texttt{CELEB-DF-V2} (as in Section~\ref{sec:impdetector}), and then evaluated them against \texttt{CELEB-M}, a specialized dataset consisting of real and deepfake images from \texttt{CELEB-DF-V1}~\cite{li2020celeb}.
The process of creating \texttt{CELEB-M} follows the same settings as the testing datasets implemented in Section~\ref{sec:impdetector}.
As deepfake images in this dataset were synthesized using the same method as those in the detectors' training set, the intention was that a detector learning \emph{method-specific} properties in deepfakes would exhibit better generalizability on \texttt{CELEB-M}, compared with the performance on the other unseen ones.

Second, we selected the detectors trained on \texttt{F2F}, and then evaluated them against \texttt{FS-I} and \texttt{NT-I}.
Different from \texttt{FS} and \texttt{NT}, \texttt{FS-I} and \texttt{NT-I} are two specialized datasets where the real and deepfake images share the same human identities as the ones in \texttt{F2F}.
The intention was that a detector learning \emph{human identity-specific} properties in deepfakes would exhibit better generalizability on \texttt{FS-I} and \texttt{NT-I}, compared with the performance on other unseen datasets.

\emph{Intuition.} For those detectors originally showing poor generalizability, if some abnormal (good) generalizability results are observed on any specialized datasets, they must have learned certain properties that are unwanted for detectors.
Following this intuition, our experiments in RQ2 were conducted based on those detectors exhibiting the poorest generalizability (performance degradation over 30\%) in Table~\ref{tab:result1}.

\emph{Measure.}
We present the results of our evaluation on \texttt{CELEB-M} in Table~\ref{tab:result2-1}, and the ones on \texttt{FS-I} and \texttt{NT-I} in Table~\ref{tab:result2-2}.
For convenience, the two tables repeat the average accuracy and degradation on the \texttt{FS}, \texttt{NT}, \texttt{DF}, and \texttt{DFD} datasets.
They also show the average accuracy of detectors on the specialized datasets, as well as the degradation in performance when compared to the original accuracy.
In the rightmost columns in the two tables, we measure a detector's generalizability change (from unseen to specialized datasets) by computing its performance degradation difference between the two groups of datasets.
Note that for the degradation difference, a higher value indicates \emph{less degradation} (and thus better generalizability) on \texttt{CELEB-M} (or \texttt{FS-I} and \texttt{NT-I}) compared to the other unseen datasets; negative values indicate that the generalizability gets worse on the specialized datasets.

\textbf{Result Analysis.}
First, in Table~\ref{tab:result2-1}, we observe that the difference in degradation is over 10 percentage points for the top six detectors (e.g., $24.95\%$ for \texttt{MesoNet1}), suggesting that their generalizability is significantly improved when encountering the \texttt{CELEB-M} dataset.
For the other detectors at ranks \#7-9, although the differences in degradation are not remarkable, they still perform better accuracy and generalizability on \texttt{CELEB-M} compared to other unseen datasets.
For example, the rank \#7 \texttt{ShallowNet1} achieves an accuracy of 62.85\% on \texttt{CELEB-M} that is higher than the ones on \texttt{FS}, \texttt{NT}, \texttt{DF}, and \texttt{DFD} (58.05\%, 50.20\%, 54.90\% and 49.60\%).
Overall, these results indicate that using the same synthesis method to generate the unseen dataset leads to better accuracy, suggesting that detectors have learned (unwanted) method-specific properties of deepfakes.

Second, in Table~\ref{tab:result2-2}, we observe that the difference in degradation is below 10 percentage points for all the detectors (e.g., $8.94\%$ for \texttt{ShallowNet4}), indicating little-to-no impact when using the same human identities in the specialized datasets.
The bottom four detectors present negative values (e.g., -3.19\% for \texttt{MesoInception5}), and their absolute values are also below 10 percentage points, which also confirms \texttt{FS-I} and \texttt{NS-I} are not unique compared to other unseen datasets.
Overall, these results suggest that these detectors have not learned human-identity features.

In general, these results are consistent with the perception that learning properties specific to synthesis methods lead to generalization issues across datasets.
In practice, we do not want detectors to focus on such properties, but rather the intrinsic characteristics of deepfakes in general.\\

\begin{tabular}{lp{2.2in}}
	\textbf{\underline{Finding 2}:} & Detectors are learning unwanted features specific to synthesis methods, and human identities do not appear to have a significant effect on generalizability.
\end{tabular}

\subsection{Learning Discriminative Features (RQ3)}
\label{RQ3}

The above experiments for RQ2 (Section~\ref{RQ2}) suggest that detectors may be learning (unwanted) properties that are specific to particular deepfake generation methods, which may impede the ability of detectors to learn those intrinsic features across different methods.
In this RQ, we explore the other side of the coin: using interpretable AI techniques, can we find any features that detection models are learning, and may be discriminative enough to distinguish real and fake images in general?

To answer this, we applied Gradient-weighted Class Activation Mapping (Grad-CAM)~\cite{selvaraju2017grad}, a popular technique used in computer vision, helping to interpret neural networks~\cite{zhang2021survey}.
We present the background of this approach first, before presenting our following visualization and quantitative experiments and results.

\textbf{Background.}
Grad-CAM implements neural network interpretation by generating a heatmap highlighting the regions that contribute the most to the neural network's decision.
In our case, the inputs of Grad-CAM are test deepfake images and detectors, and the outputs are heatmaps that visualize the regions highly contributing to the decision of `real' or `fake', based on class prediction scores.

Specifically, given a detector and a desired class (e.g., `fake'), an image input is first propagated through the detector's Conv layers and then through task-specific computations to obtain a raw score for the class.
The gradients are set to 1 for the desired class and 0 for the other.
Then, the signal is backpropagated to the rectified convolutional feature maps, which are combined to compute a coarse location heatmap (i.e., the detector makes particular decisions based on the coarse location).
Finally, the coarse heatmap is pointwise-multiplied using guided backpropagation to obtain the final heatmap.

\textbf{Visualization Experiments and Results.}
We first conduct a visualization experiment to preliminarily observe the features for deepfake image detection.
Specifically, we implemented Grad-CAM based on the Tf-Explain project~\cite{tfexplain2021} and applied it to our deepfake detectors (Section~\ref{sec:impdetector}) on the unseen \texttt{FS}, \texttt{NT}, \texttt{DF}, and \texttt{DFD} datasets.
We used the settings recommended in~\cite{selvaraju2017grad}, i.e., to visualize the contributing features in the last convolutional layer of each detector, since such layers are close to output layers and involve relatively complete features for classification.
Moreover, instead of superimposing heatmaps on the original images, we generated pure heatmaps based on the applyColorMap function in OpenCV~\cite{bradski2000opencv} to facilitate the analysis of visualized features.

\begin{figure}[!t]
	\hfill
	\begin{center}
		\includegraphics[width=\linewidth]{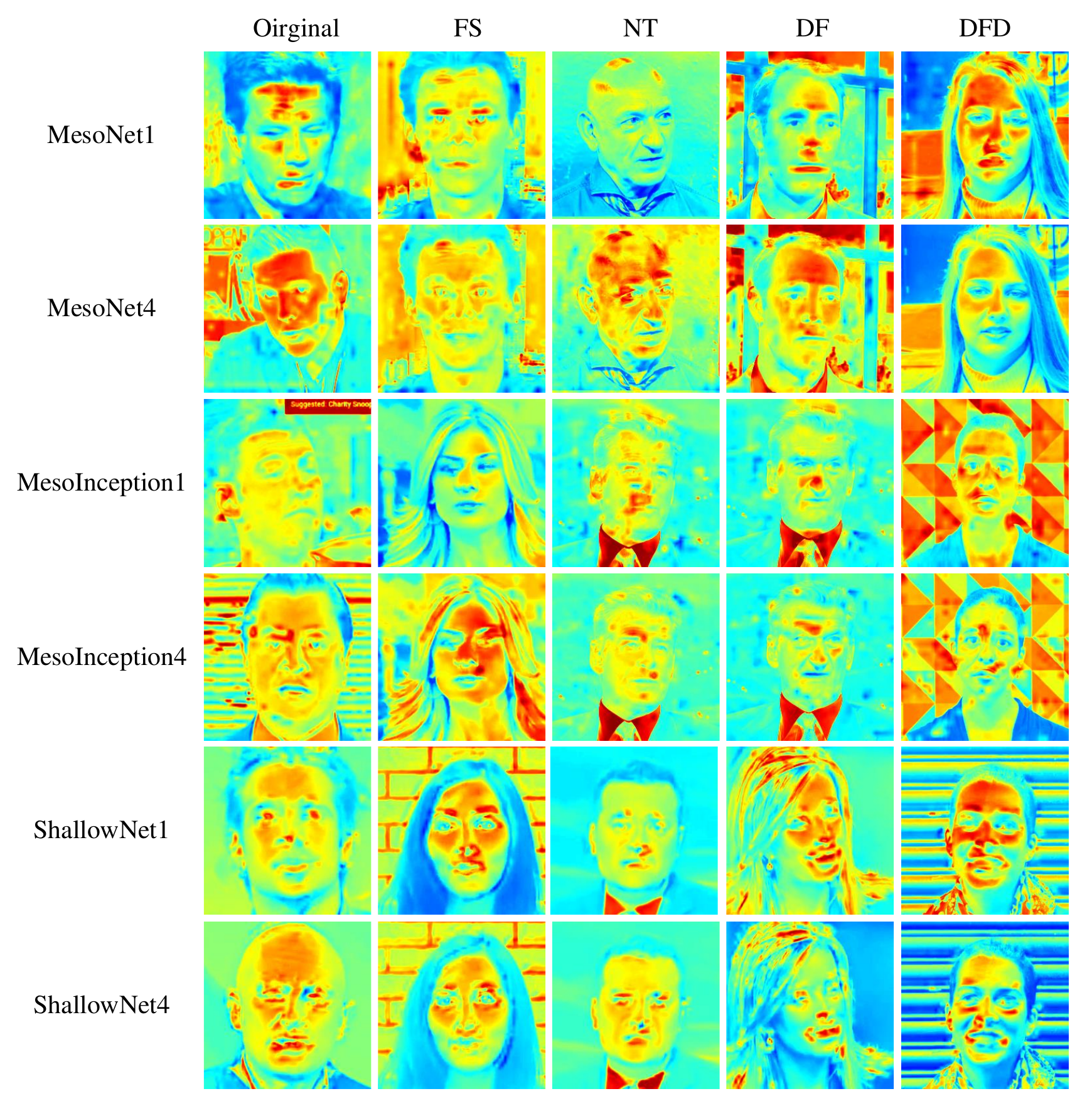}
	\end{center}
	\caption{Heatmaps visualizing the regions contributing to detectors' classification decisions across datasets: the intense and cool colors (such as red and blue) indicate high and low contributions respectively.}
	\label{fig:result3-1}
\end{figure}

\emph{Result analysis.}
Figure~\ref{fig:result3-1} presents a few example feature visualization results, where colors of higher intensity (such as red) indicate higher contributions towards the detectors' classification decision.
Ideally, the contributing features would be located in certain fixed facial regions, suggesting that detectors have extracted features that discriminate between real and fake in human faces.
However, as illustrated in the figure, the visualized features are scattered across different regions of the deepfakes, including foreheads, eyes, mouths, and even backgrounds, suggesting that they are not consistently extracting discriminative features.

\textbf{Quantitative Experiments and Results.}
We would like to provide a preliminary exploration through the above visualization results, and then conduct quantitative experiments to further explore discriminative features.

\emph{Intuition.}
We quantitatively measured the similarity between heatmaps generated by detectors that have the same architectures and training sets (e.g., \texttt{MesoNet1}, \texttt{MesoNet2}, \texttt{MesoNet3}).
That is, these detectors are `congenetic', i.e., they derive from the same original detector and are constructed with certain model augmentations.
Ideally, given a test deepfake, such detectors could focus on similar features on it, otherwise, their extracted features could be `unrestrained' and are not discriminative enough for classification.

\begin{figure}[!t]
	\hfill
	\begin{center}
		\includegraphics[width=\linewidth]{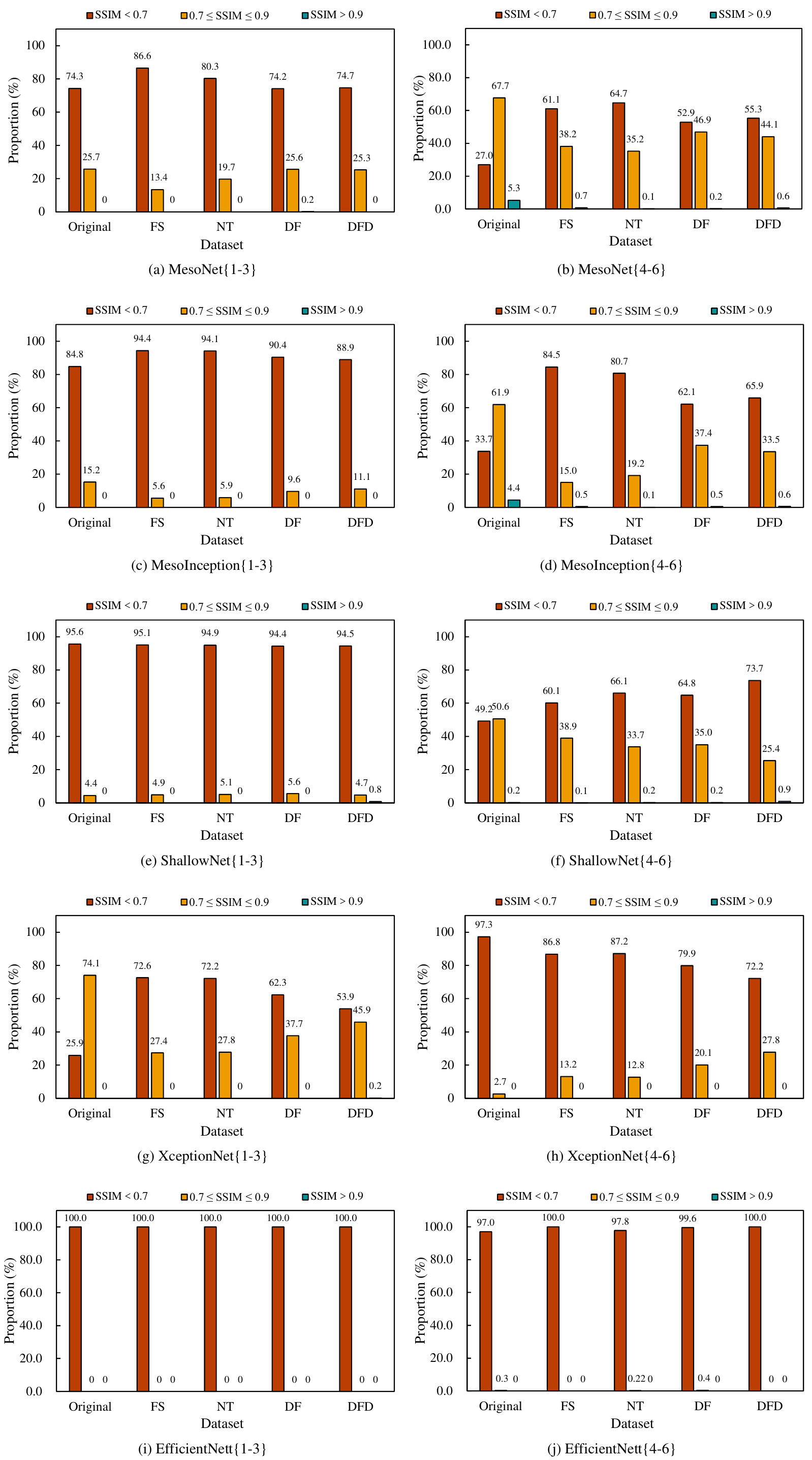}
	\end{center}
	\caption{Similarity (SSIM) of heatmaps generated by groups of detectors on unseen datasets.}
	\label{fig:result3-2}
\end{figure}

In particular, we applied Structural Similarity (SSIM)~\cite{wang2004image} for this purpose. 
SSIM evaluates the similarity between images based on three aspects~\cite{wang2004image}, i.e., luminance, contrast and structure. 
These values are combined into a similarity score ranging from 0 to 1, where 1 indicates identical images. 
SSIM is selected since luminance and contrast are sensitive to the brightness of images, which matches the characteristics of the heatmaps of deepfakes.
For each group of detectors, we applied them to 1000 deepfakes from each of the unseen datasets, generated heatmaps accordingly, and measured their SSIM.
Usually, 0.9 is applied as an SSIM threshold to determine whether test images have high similarity; 0.7 is the threshold to determine whether test images are dissimilar~\cite{wang2004image}; and values between 0.7 and 0.9 indicate low amounts of similarity.

\emph{Result analysis.}
We present our quantitative results in Figure~\ref{fig:result3-2}.
In the figure, we observe that the SSIM of most groups of heatmaps is below 0.7 (the red columns) indicating that most heatmaps are dissimilar, despite generating them in groups of detectors with similar underlying architectures.
As exceptions, a few groups of heatmaps show higher SSIM values (between 0.7 and 0.9; the orange columns) indicating low amounts of similarity.
For example, in Figure~\ref{fig:result3-2}~(b), $66.7\%$ of the heatmaps generated for the original testing set (\texttt{F2F}) have SSIM values between 0.7 and 0.9.
Such values are reasonable in the original testing set where the extracted features are likely to be more similar than those in unseen ones.
Nonetheless, across the ten histograms, very few groups of histograms achieved SSIM values above 0.9 (the green columns).
Recall that these `congenetic' detectors should have extracted similar features from the same testing deepfakes, and the contrast results confirm that the contributing features are unrestrained, other than being discriminative for generalization.\\

\begin{tabular}{lp{2.2in}}
	\textbf{\underline{Finding 3}:} & Detectors are extracting unrestrained features that are not discriminative enough for generalizable detection.
\end{tabular}

\subsection{Contributing Neurons (RQ4)}
\label{RQ4}

The results thus far suggest a fundamental challenge must be overcome, before generalizable (zero-shot) deepfake detection can become a realistic prospect.
In this RQ, we approach the issue of generalizability from a more granular perspective, asking whether detection models have \emph{universally contributing neurons} across seen and unseen datasets.
Rather than exploring \emph{features} from input data as in RQ2 and RQ3, contributing neurons may facilitate a new way to identify features for classification.
If such neurons exist, they may eventually provide a foundation for building generalizable detectors.

In a neural network, some neurons contribute more to classification results than others, a fact that has been exploited in other work, e.g., neural network repair~\cite{sun2022causality}.
Inspired by this work, we propose a \emph{causality-based} method to analyze detection models, so as to identify contributing neurons that are causally associated with classifying a testing image as real or fake.
We present the background of this approach first, before presenting the details of our experiments and results.

\textbf{Background.}
We identify the contributing neurons by interpreting a detector as a Structural Causal Model (SCM), which is a significant causality analysis model that has been applied to multiple neural network interpretation tasks, e.g., \cite{chattopadhyay2019neural, zhao2021causal}.
Formally, an SCM is defined as follows~\cite{pearl2009causality}:

\begin{definition}[Structural Causal Model (SCM)]
	An SCM is a four-tuple $C(Y,K,f,P_k)$, where $Y$ and $K$ are finite sets of endogenous and exogenous variables respectively, $f$ is a set of functions $\{f_1,f_2,\ldots,f_n\}$, and $P_k$ is a probability distribution over $K$.
	Each function of $f$ represents a causal mechanism such that $\forall y \in Y, y_i = f(Pa(y_i),k_i)$, where $Pa(y_i)$ is a subset of $Y \backslash \{y_i\}$ and $k_i \in K$.
\end{definition}

Intuitively, an SCM can be thought of representing causal relationships in a system as a directed acyclic graph~(DAG), where variables are represented as nodes and causal relationships between them are represented as edges. 
Variables are distinguished as endogenous or exogenous based on whether or not (respectively) they are affected by other variables in the system.

We interpret a neural network-based deepfake image detector as an SCM to analyze the relationship between its neurons and predictions.

\begin{definition}[Detection Model as SCM]
	A deepfake detection model $N(y_1,y_2,\ldots,y_n)$ is interpreted as an SCM by the four-tuple $C([y_1,y_2,\ldots,y_n],K,[f_1,f_2,\ldots,f_n],P_k)$, where $y_i$ is the set of network neurons at layer $i$, $y_1$ and $y_i$ are the neurons at input and output layers respectively, $K$ is a set of exogenous random variables to serve as causal factors for input-layer neurons $x_1$, and $P_k$ is a probability distribution over $K$.
	Corresponding to each $y_i$, $f_i$ is the set of causal functions for the neurons at layer $i$.
\end{definition}

\begin{figure}[t]
	\hfill
	\begin{center}
		\includegraphics[width=0.8\linewidth]{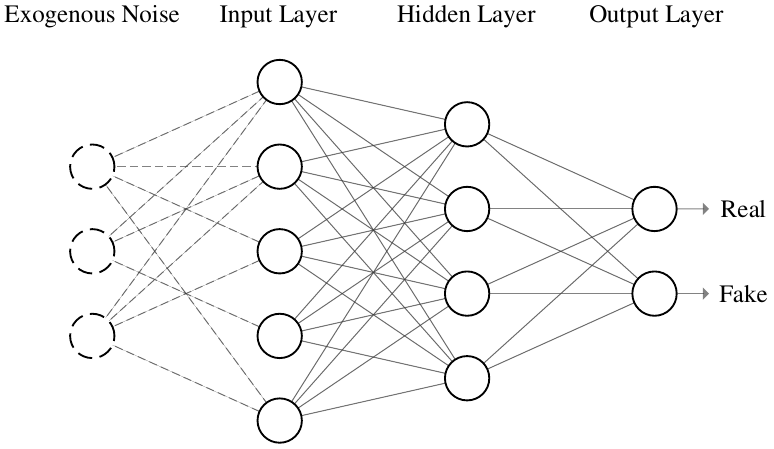}
	\end{center}
	\caption{Interpretation of a detection network as an SCM.}
	\label{fig:scm}
\end{figure}

Figure~\ref{fig:scm} illustrates a deepfake detection model to be interpreted as an SCM.
In the figure, the solid-line neurons correspond to endogenous variables, whereas the dotted circles represent the latent exogenous noise variables which serve as causal factors for input neurons.
Functions connect the layers, with edge weights corresponding to cause-effect relations.
Note that the input-layer neurons can be denoted as the functions of the latent exogenous variables, and are assumed not causally related to each other (but can be jointly affected by latent noise).

\begin{definition}[Average Causal Effect (ACE)]
	Let $y$ denote a neuron of a neural network-based detector $N$ and $z$ denote the measure of $N$'s degradation (as in Sections~\ref{RQ1}).
	Then, the ACE of $y$ on $z$ is:
	\begin{equation*}
		\label{ace}
		ACE_{do(y=\gamma)}^{z}
		=\mathbb{E}[z|do(y=\gamma)]
		=\int_{z}zp(z|do(y=\gamma))dz,
	\end{equation*}
	where $do(\cdot)$ is the corresponding interventional distribution defined by SCM,
	and $ACE_{do(y=\gamma)}^{z}$ is calculated by the interventional exception of $z$ given $do(y=\gamma)$.
\end{definition}

Specifically, given a detector $N$, we sample input $x$ from an unseen dataset $U$ and denote $N_{ac}(x)$ as $N$'s predicted class corresponding to $x$'s ground-truth label $L(x)$.
For each neuron $y$, we keep $y=\gamma$ and measure its contribution to $z$ by calculating
$ACE_{do(y=\gamma)}^{N_{ac}(x)}$ with the sampled inputs.
Ultimately, by calculating the causal effect of each neuron in $N$ (based on the testing samples in $U$), i.e., their contributions to $N$'s generalization, the contributing neurons are identified.

\begin{figure*}[!t]
	\hfill
	\begin{center}
		\includegraphics[width=\linewidth]{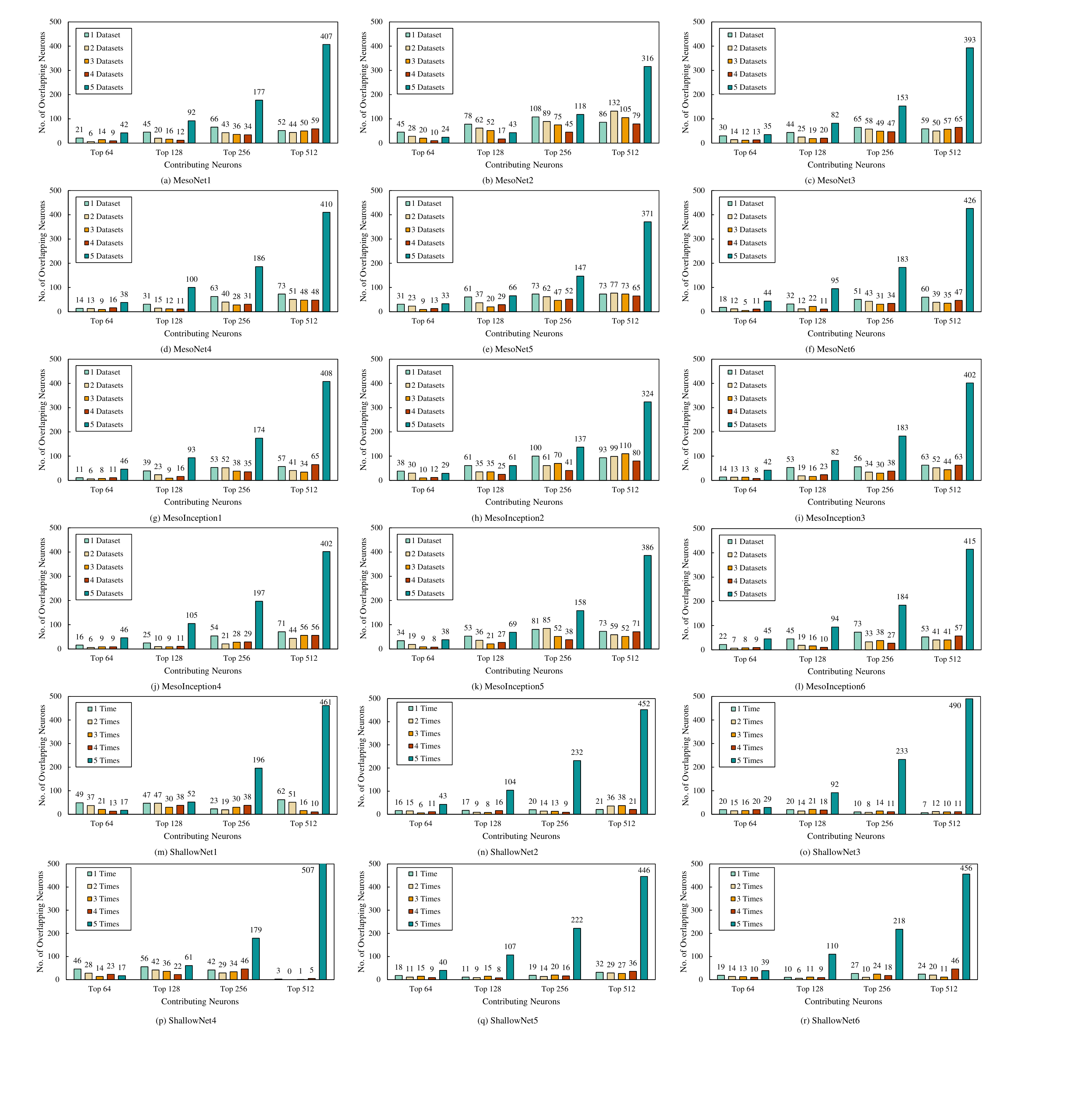}
	\end{center}
	\caption{Causality analysis: the number of (overlapping) neurons that have contributed to classification across the five testing datasets (\texttt{CELEBV2}, \texttt{FS}, \texttt{NT}, \texttt{DF}, and \texttt{DFD}).}
	\label{fig:result4}
\end{figure*}

\textbf{Experiments.}
Based on the above, we performed causality analysis on the detectors implemented in Section~\ref{sec:impdetector}, in order to identify neurons contributing to both seen and unseen datasets (and thus may potentially be involved with generalizable features).
Specifically, we identify the contributing neurons in the first dense layer of the \texttt{MesoNet}, \texttt{MesoInception}, and \texttt{ShallowNet} detectors, since such layers receive significant features extracted by Conv modules and conduct actual classification functions.
\texttt{XceptionNet} and \texttt{EfficientNet} are not selected since they both contain only one dense layer that is augmented to perform binary classification, and thus their dense layers contain limited numbers of neurons that are not enough to analyze and compare contributions (We will further discuss the limitations on unselected detectors in Section~\ref{threats}).

In our experiment, based on the five datasets containing the detectors' original testing ones as well as the four unseen datasets, we identified the top 64, 128, 256, and 512 contributing neurons in the detectors' first dense layers (1024 neurons in total; thus the top 1/16, 1/8, 1/4, and 1/2 respectively).
Following the causality analysis, we then counted the number of `overlapping' neurons (i.e., the contributing neurons involved across multiple datasets) and the number of datasets they contributed to a decision, up to a total of 5 times (\texttt{CELEBV2}, \texttt{FS}, \texttt{NT}, \texttt{DF}, and \texttt{DFD}).

\textbf{Result Analysis.}
We present our results in Figure~\ref{fig:result4}, where each histogram presents the outcome of a causality analysis on one detector.
For example, in Figure~\ref{fig:result4}~(a), when identifying the top 64 contributing neurons of \texttt{MesoNet1}, it is observed that 42 of them contributed to classification for all five testing datasets (\texttt{CELEBV2}, \texttt{FS}, \texttt{NT}, \texttt{DF}, and \texttt{DFD}), whereas, for example, 21 of them only contributed to the decision for one dataset.

As can be seen in the green columns in these histograms, we observe that neurons are most frequently contributing to \emph{all five datasets}, regardless of whether considering the top 64, 128, 256, or 512 neurons.
This was especially the case for the last category (i.e., in the top 512 neurons).
For example, when analyzing \texttt{MesoNet1} in Figure~\ref{fig:result4}~(a), 407 neurons (out of the top 512) were identified as contributing to the decision across all five datasets.
The exception cases are shown in Figures~\ref{fig:result4}~(m) and (p): when identifying the top-64 and top-128 contributing neurons in the \texttt{ShallowNet} detectors, we do not observe a significant number of neurons contributing across all five datasets in comparison to the other frequencies.
However, when considering the top 256 or 512 contributing neurons, their results are consistent with other detectors.
Such results are reasonable since detectors might make unstable decisions based on the features involved in the limited 64 or 128 neurons.
In general, the results suggest that there are neurons that are universally contributing to the detection of both the original and unseen testing datasets, which may provide a useful basis for building generalizable detectors in the future.

Contrast this with our observations in Section~\ref{RQ3} (RQ3), in which we found that detectors extracted unrestrained features from inputs.
Our causality-based analysis took a more granular view of neural networks, and observed that it is possible to identify neurons that contribute to the decision across all five unseen datasets.
Thus, we conjecture that detectors focusing on these universally contributing neurons may be more successful at generalizing to different datasets.

\emph{Discussion.}
Existing efforts have provided some inspiration for making use of important neurons in neural networks, such as pruning those low-contributing neurons or enhancing models based on transfer learning~\cite{molchanov2016pruning}. 
However, applying these strategies to deepfake detection still has a long way to go, since the variety of deepfake datasets can lead to various potential risks, such as overfitting, catastrophic forgetting~\cite{kemker2018measuring}, or performance degradation due to the reduction of network parameters.
Despite this, compared with existing generalization methods (Section~\ref{sec:generalizability}) that mainly focus on fine-tuning detectors at layer levels, our results suggest a granular neuron-level perspective may have greater potential for achieving generalizable detection.\\

\begin{tabular}{lp{2.2in}}
	\textbf{\underline{Finding 4}:} & Detectors have universally contributing neurons across seen and unseen datasets.
\end{tabular}

\section{Research Directions}
\label{discussion}

In this section, we discuss and recommend potential research directions based on the findings and conclusions from our empirical study.

\emph{Building deepfake detectors based on causality analysis.}
Based on our studies in Section~\ref{RQ4} (RQ4), through causality analysis, we can identify the contributing neurons that may involve generalizable features.
Thus, a prospective research direction is to build detectors based on those universal contributing neurons.
For example, when training detectors, we might increase the weights of the universally contributing neuron-related parameters, or freeze the remaining low-contributing neurons. 
As we discussed in the concluding paragraph of Section~\ref{RQ4} (RQ4), applying causality analysis to achieve generalizable detection still has a long way to go.
Fortunately, we have taken the first step of demonstrating the existence of these universally contributing neurons, the use of which is still worth exploring in further research.

\emph{Preventing detectors from learning interferential features.}
Based on our findings in Section~\ref{RQ2} (RQ2) and Section~\ref{RQ3} (RQ3), deepfake detectors have learned unwanted synthesis-method properties, and extract unrestrained features that are not discriminative enough for generalization.
These findings indicate a potential research direction, i.e., applying certain measures to prevent detectors from learning interferential properties, e.g., applying \emph{Dropout}~\cite{srivastava2014dropout} operations supplementing regularization terms in loss functions~\cite{ying2019overview}, so as to enable detectors to focus on those intrinsic features.

\emph{Evaluating future deepfake detectors.}
To perform our study, we created a repository of deepfake datasets as well as detectors in multiple settings.
Recall our studies in Section~\ref{RQ1} (RQ1), the generalization problem is common across detectors.
Therefore, our implemented datasets and detectors can serve as baselines to evaluate whether an upcoming deepfake detector is reliably generalizable.

\section{Threats to Validity}
\label{threats}

In this section, we discuss potential threats to the validity of our empirical study.

\emph{External validity: limitation of the proposed causality analysis method.}
In Section~\ref{RQ4} (RQ4), we performed causality analyses on the dense layers of \texttt{MesoNet}, \texttt{MesoInception}, and \texttt{ShallowNet}.
Although our proposed method applies to multiple detectors, it is not enough to analyze those detectors with only one or without any dense layers, such as \texttt{XceptionNet} and \texttt{EfficientNet}, since such layers contain limited neurons to perform binary classification and it makes no sense to analyze such neurons.
Despite this, our aim in RQ4 is to explore the possibility towards generalizable detection, and the potential neurons have been found from detectors.
Thus, although the proposed causality analysis method could still be improved, our existing explorations meet the goal of this study.

\emph{Internal validity: selection of deepfake detectors.}
We have selected multiple state-of-the-art deep learning-based detectors (as in Section~\ref{sec:detector}).
Although visual feature-based detectors also report some promising results (mainly on seen datasets), they are not suitable for generalizability studies due to their instability (limited by objective factors).
Moreover, such detectors are still in the early stages of being applied in practice.
Our selected deep learning detectors are most likely to generalize and are thus most suitable for this study.

\emph{Construct validity: implementation of deepfake detectors.}
We have trained multiple detectors (as in Section~\ref{sec:impdetector}) whose performance can potentially be affected by their implementations, such as the optimizer or learning rates.
To mitigate such concerns, we implemented these detectors based on the settings that have been evaluated by their original authors or related effective implementations.
Moreover, we have evaluated these detectors and observed promising original accuracy, and thus they meet the requirements to perform this study.

\section{Conclusion}
\label{conculsion}

In this work, we conducted an empirical study to understand how generalizable deepfake image detectors are.
Based on our study, we have: (1)~confirmed that existing detectors lack generalizability, especially in zero-shot settings;
(2)~found that detectors learn unwanted method-specific and unrestrained features that are not discriminative for generalization; and (3)~proposed a causality analysis method that identifies the presence of universally contributing neurons across unseen datasets, which may open a path forward to solving the generalizability problem.
Finally, we have provided a repository of datasets and detectors, which we believe can provide baselines for evaluating the generalizability of new deepfake detection approaches.

\bibliographystyle{plain}
\bibliography{sample}

\end{document}